# ARTIFICIAL NEURAL NETWORK APPROACH FOR

# CONDITION-BASED MAINTENANCE

Mostafa Sayyed

Master of Engineering Management

Universiti Putra Malaysia

Abstract

In this research, computerized maintenance management will be investigated. The rise of maintenance cost forced the research community to look for more effective ways to schedule maintenance operations. Using computerized models to come up with optimal maintenance policy has led to better equipment utilization and lower costs. This research adopts Condition-Based Maintenance model where the maintenance decision is generated based on equipment conditions. Artificial Neural Network technique is proposed to capture and analyze equipment condition signals which lead to higher level of knowledge gathering. This knowledge is used to accurately estimate equipment failure time. Based on these estimations, an optimal maintenance management policy can be achieved.

Abstrak

Dalam kajian ini, pengurusan penyenggaraan berkomputer akan disiasat. Kenaikan kos penyelenggaraan terpaksa komuniti penyelidikan untuk mencari cara yang lebih berkesan untuk menjadualkan operasi penyelenggaraan. Menggunakan model berkomputer untuk datang dengan dasar penyelenggaraan optimum telah membawa kepada penggunaan peralatan yang lebih baik dan kos yang lebih rendah. Penyelidikan ini menggunakan model Penyelenggaraan Berasaskan Keadaan di mana keputusan penyelenggaraan dihasilkan berdasarkan keadaan peralatan. Teknik Rangkaian Neural Buatan adalah dicadangkan untuk menangkap dan menganalisis isyarat keadaan peralatan yang membawa kepada tahap yang lebih tinggi daripada pengumpulan pengetahuan. Pengetahuan ini digunakan untuk menganggarkan dengan tepat masa kegagalan peralatan. Berdasarkan anggaran ini, dasar pengurusan penyelenggaraan optimum dapat dicapai.

## Acknowledgement

In the name of Allah, Most Gracious, Most Merciful Who had bestowed me knowledge, patience and strength in completing this project report.

Alhamdulillah, I want to express my most heartfelt gratitude and appreciation to my supervisor, Dr …… for his guidance and patience throughout the whole development of this work. His advices were very encouraging and help me to rediscover my confidence to complete this project. I am very grateful for everything that he shared to me and lead me to perform better. Also, I would to give immense thanks to my examiner, Dr….. for his encourage comments and ideas.

Next, I would like to express my highest gratitude to Abu Dhabi police who have been supporting me in my studies as well as inspiring me till I am able to achieve to this level of education.

Finally, I would like to thanks to all my colleagues who have helped to give me full support and motivation and not to forget my brother and sister.

Thank you.

Mostafa Sayyed

Approval

This project report is submitted to the Department of Mechanical and Manufacturing Engineering, Faculty of Engineering, Universiti Putra Malaysia and has been accepted as a partial fulfillment of the requirement for the degree of Master of Engineering Management. The members of the examination Committee are as follows:

Supervisor

Dr …….

Department of Mechanical and Manufacturing Engineering

Faculty of Engineering

Universiti Putra Malaysia

Examiner

Dr ………….

Department of Mechanical and Manufacturing Engineering

Faculty of Engineering

Universiti Putra Malaysia

Declaration

I hereby declare that the project report is my original work except for quotations and citations, which have been duly acknowledged. I also declare that it has not been previously, and is not concurrently, submitted for any other degree at Universiti Putra Malaysia or at any other institution.

___________________________________

Mostafa Sayyed

GS

Date:

# Table of Contents

List of figures

Chapter 1:     Introduction

The emergence and growth of computers has brought technology that has eliminated the need to have manual labor in certain roles and processes in organizations. The development of computerized management software has made maintenance work easier for organizations (Hernandez, 2001, 06). Over the years, these management systems have been developed to include more functions and make them more efficient in the roles. This has led to the number of operations in an organization that are under the management of a computerized system more, making work easier and in the process reducing the number of employees in these organizations (Idhammar, 1992).

The computerized management systems have resulted in many benefits to organizations that have adopted them. Most organizations have reported a 28.3% increase in maintenance productivity, 20.1% reduction in equipment's downtime, 19.4% savings in lower material costs, 17.8% reduction in maintenance, repairs operation (MRO) inventory, and 14.5 months average payback time (Morton, 2009). These benefits have made the computerized management systems popular and in high demand. Competitiveness in the business world has forced organizations to adopt measures that lead to the reduced costs of operations and increment in revenue. These systems have proved reliable and managed to satisfy the needs of organizations that have adopted them. Human labor has been replaced with the computerized systems that have proven more reliable and efficient than human beings. These systems have a number of capabilities that include: the ability to monitor operations that comprise processes, equipment's state, scheduled and unscheduled process, development of reports and their storage. They track employees' performance and provide data that is important in the improvement of service delivery and utilization of the available resources. These capabilities have made them a must have for large organizations and medium sized ones, especially those involved in production and service delivery (Norman, 1997).

The need to have more efficient management systems is high. This is a result of the need to have systems that incorporate more functions, are more user-friendly and highly efficient. Just like any other technology, there are changes that are taking place in the designs of the computerized management systems. This is supposed to overcome some of the problems that organizations that adopt these systems encounter. Some of the management systems in the market are not user-friendly. This has created a problem when the staff takes more time to understand how to operate and integrate the system into the organization (Patton, 2007). This has resulted in hitches when organizations operating them fail to understand how to respond to the system demands. Also, there is a need to design highly efficient systems that can handle more tasks than the existing ones. This is to suit the demands of the market where organizations want to replace human labor with the computerized systems to cut costs, increase efficiency, and improve on other aspects of their operations that will result in reduction in costs (Raouf, Ali & Duffuaa, 1993).

The existing designs have various capabilities that can be improved to make them more efficient. They may include other functions that are lacking in the existing models (Autin, 1998). This project will be concerned with the analysis of the existing designs and the design of a system that has features lacking in the current models. This will result in a better system serving more functions than the available ones do.

1.1     Problem Statement

The usual maintenance policy in practice is to wait for equipment failure before performing any maintenance action. It turned out that this policy causes huge losses in term of Down-Time cost. In the last two decades, Preventive Maintenance (PM) gain popularity due to its cost effective approach. PM tries to perform maintenance before equipment failure so that no down-time is experienced and full equipment's utilization is achieved. However, performing maintenance on equipment long before its failure time introduces unnecessary cost; especially, if maintenance is being performed frequently on stable equipment.

Condition-Based Maintenance (CBM) has the ability to tackle this challenge by estimating the remaining time until equipment failure based on equipment conditions. These conditions can be any feature of equipment which indicates unstable behavior such as loud sound or high temperature. Researchers have been developing CBM models to accurately estimate failure time. All of their models are based on statistical approach to estimate failure time which leads to low accuracy. This problem is due to the fact that statistical approach usually depends on couple of numbers. For example, the average (mean) or standard deviation of equipment temperature does not tell us anything about how fast the temperature is changing which is very important information that may help to increase the accuracy of failure time estimation. Therefore, adopting statistical approach leads to leaving so much of valuable information and focusing on couple of numbers which problematic and not wise.

## 1.2 Objectives of the Study

The aim of this study is to develop Condition-Based Maintenance mechanism that has the ability to capture all necessary information to accurately estimate equipment failure time. The following are research objectives:

I. To design a policy for identifying any relevant information of equipment conditions and behavior that can be used to enhance failure time estimation.

II. To develop neural network based mechanism that approximate equipment failure time by extract the nonlinear hidden relationship among equipment's and components locally and globally.

III. To design maintenance management policy that takes advantage of neural failure time estimator to achieve the lowest cost and highest utilization.

## 1.3 Scope of the Study

The need to have computer management systems that can do more than the current ones have led to the desire to analyze the existing ones and identify the areas that can be improved to make better systems. There are emerging issues that can be resolved by making some improvements in the existing designs so as to make better systems that can store records of all existing assets of an organization, as well as track and keep construction information of a building or product using the Building Information Modeling (BIM). This information can be used to commission facilities and validate performance (Modern Machine Shop, 2011). These two features that are on top of the available ones would make these systems better than they are.

This project aims at understanding the way how CMMS systems work and how they may be modified to include the above-named features that are important but lacking in the current designs. Analysis of the existing systems will be done, and a new design will be formulated to tackle the emerging issues.

Research on the computerized maintenance systems is of great benefit to the current and future organizations. The need to improve efficiency and be able to utilize available resources has to be done using technology that is improving rapidly. The government agencies that have been in operation for many decades before information technology was adopted often encounter problems when trying to use their available assets (Allen, 1999). This problem has often led to underutilization of the available resources. The need to rate buildings and products based on their quality on production is usually high to ensure that they are rated accordingly and where necessary discarded. This information can be stored in the computerized maintenance systems and used later when the building or system is in use (Elliott, 2000).

The improved efficiency and proper utilization of the available resources and assets to a company would reduce wastage and cost of production, leading to lower costs of products they sell and higher profits for the organizations (Modern Machine Shop, 2011). Higher quality is another likely benefit of the high quality CMMS systems. This makes it of paramount importance to engage in research to improve the existing designs and add features that will make them better. Changes in technology support better systems, and the availability of information will be an added advantage in the development and use of the systems.

Improvement in features of the computerized maintenance systems and their user-friendliness will make it easy for people to adopt these systems and use them, especially those in the developing countries where information technology is still at its infancy in terms of adoption (Morton, 2009). These improvements will enable organizations that adopt these systems obtain the full benefits of technology.

Chapter 2:    Literature Review

This chapter analyses the information regarding the computer maintenance management systems that are available. This information is derived from journals and books that contain the researches that have been conducted in this field. The researches published provide details of the research approach used, the methodology adopted, the results obtained, recommendations, and conclusions. These journals are peer reviewed, making them acceptable in academics as sources of information. These journals contain information that is relevant to this research study.

## 2.1    Computerization

Computers have led to the evolution of technology to such an extent that they have taken over some of the duties that were undertaken by human beings. Computers have proven to be more efficient than a human being (Raouf, Ali, & Duffuaa, 1993). This has led to the development of software that enable computers play complex roles, thus eliminating lots of human labor and errors that come with it. Their evolution from the simple office computers that can be used to type and save documents and play a few media files has coincided with the need to reduce the cost of production and improve the entire production system in the manufacturing industry. This has been a result of the need to increase the efficiency and maintain the quality of good produced (Sahoo & Liyanage, 2008). Human beings have been playing this role for centuries, but they have always messed up due to human errors. High competition in the marketing world and the need to reap maximum profits without having to increase the prices have necessitated the need to cut the cost of production and improve quality to attract more customers (Raouf, Ali, & Duffuaa, 1993). Consequently, manufacturers and other business people have been forced to rely on technology to achieve this target. The ability of computers to play myriad roles with a high efficiency level and a low operation cost has led to their popularity. Computers have taken over in most companies playing many roles that human beings used to undertake. In addition, computers record the data as programmed and based on the security features of the system; thus, the data is safe unlike when human beings used to record and file it (Sahoo & Liyanage, 2008). Computers have solved issues that concern the efficiency. Also, they have created a reliable system that can undertake many duties in a plant based on the program that it is running on is able to do (Raouf, Ali, & Duffuaa, 1993). Technology is evolving at a fast rate developing different software to perform different roles. All these programs are relying on the capabilities of the computer to do complex and easy functions efficiently and easily. One of the systems that have been developed to use a computer is the computer maintenance management system.

## 2.2    Maintenance Management

Maintenance plays a key role in organizations especially that deal with manufacturing of goods. Machines need to be maintained after certain duration of time to ensure that they are functioning well. The task of maintenance has been carried by human beings for many years. This has been done by monitoring the productivity of machines and establishing the time within which the machines are serviced to ensure they are capable of maintaining their productivity levels. This has worked though not to the efficiency levels desired, thus creating the need to have systems that can do that (Morton, 2009). The main reason is that it is cumbersome for human beings to monitor the production of a machine and to track changes that may signify an imminent breakdown. In most cases, human beings have noted problems with machines if they stall during their work. This is solved by repairing the machine. The problem may also be noted if an observation is made, showing a decline in production levels of the machine. This may be a result of a problem that has been developing in the machine for a few days, but it has not been detected because it had no effect on the production process. It is noted that it may have escalated to the levels of slowing down and affecting the entire process based on how vital the machine is in the factory (Idhammar, 1992). Conversely, the organization incurs huge losses that could have been avoided if the problem had been identified in the very moment it occurred. The statistical data pertaining to the machines and the whole production process is vital to determine which changes are necessary (Patton, 2007). This has always posed a problem to human beings due to their inability to memorize the whole data. A computer has proven it can handle data, thus making it an ideal equipment to monitor the production process. The abilities of the computer and the shortcomings of human beings have created the need for a computer system that can monitor the production process and identify most of these problems that human beings could not detect (Autonomous maintenance systems, 1993).

## 2.3    Computer Maintenance Management Systems

Computer maintenance management systems are computer systems that are developed to undertake maintenance management in organizations. The roles a system can play are always determined by the developer of the system (Idhammar, 1992). There are a lot of vendors of these systems in the world. Most of these systems can perform many functions that are expected from a computerized system. They have gained popularity since they were introduced into the market with most companies adopting them. Their adoption has always led to unemployment with many people getting displaced by these systems. Consequently, the cost of labor has reduced, thus leading to a decrease in the production cost (Key impacts and benefits of computerized maintenance management,, 1999). The systems have been developed to solve problems that human beings failed to solve when managing the maintenance process. Management systems have been developed with myriad abilities that enable them to multitask and efficiently solve a number of key tasks in an organization (CMMS: What you need to know, 1994). These systems are installed by organizations that try to undertake effective maintenance management by evaluating and monitoring the ongoing production processes, facilities, stores and inventory control, purchasing operation, accounting and information systems, and the entire maintenance department in the organization. This requires setting up a system that can handle huge loads of data and analyze (Eagle's ProTeus computerized maintenance management system supports global solutions, 2002). This data helps the management determine when to undertake the necessary maintenance of the facility and the processes that are undertaken there.

Traditional maintenance programs that rely on human beings identifying problems do so only when an equipment breaks down. Further problems arise because the spare parts must be ordered if none is available in the store (Idhammar, 1992). This is because the individuals responsible for keeping a record of the available spare parts may have failed to order for a replenishment of the spare parts after they run out of stock. A computerized maintenance system keeps a record of the repair dates based on the recommendations of the manufacturers of the equipment or the maintenance department (Brown & Paine, 1992). It keeps a record of the spare parts and provides reminders to ensure that the spare parts are bought in advance. The system is also repaired before it breaks down, which saves an organization lots of money (Wimsatt, 1998). This is an aspect of computerized maintenance management that has provided huge benefits and compelled many organizations to install these systems. There is a provision to include information regarding the availability of the spare parts in the spare parts inventory. This comes in handy to the maintenance department. It ensures that the spare parts are available in time and that an equipment repair will not be delayed due to the lack of the spare parts that will most certainly halt operations in the production line (H: Maintenance management, 2004). Planned maintenance has the effect of controlling costs because the repairs that are done at the right time can be budgeted and their cost can be controlled unlike those that are haphazard and dependent on machines breaking down (Antonacci, 1994). It is impossible to determine how many times a machine will break down; therefore, budgeting for repairs and maintenance cannot be made. This is the aspect of computerized maintenance management systems that have enabled them to serve the direst needs of the maintenance departments (Idhammar, 1992). Maintenance plays a key role in any organization, especially those involved in manufacturing of goods. The properly maintained machines and equipment are capable of working efficiently meeting their productions targets and ensuring that an organization is able to meet its obligations to its buyers.

2.4     Selecting the Right CMMS

Organizations that install the computerized management systems must ensure that they have made their objectives correctly to avoid installing a system that may not help them. This problem arises when an organization is not ready for the changes that the system is going to introduce. Organizations that have successfully installed these systems have initiated changes by beginning with the individuals responsible for the installation (Silverberg, & Taylor, 1999). An organization should undertake a thorough feasibility study and identify the tasks that the system they intend to install should play. This should be followed by identifying the right system in the market from a trusted vendor. This is a key role because the right vendor will provide a system whose capabilities are tested and proven in the market and also offer orientation and maintenance services (Carroll & Wilmot, 2003). The main motivation for the adoption of a computerized system is a reduction of operation costs. This is achieved through the increase of the plant efficiency, prevention of plant failures and unplanned downtime, and attainment of the high safety standards. This process of identification of a proper system requires the individuals responsible for the purchase and installation of the system to be proactive (Allen, 1999). This is important because it helps them identify issues that will arise later and that will require being addressed. This enables them to identify a system that has more features than they need. A failure to do the following results in the purchase of a system that handles the existing functions only to lose value once the organizations expands it functions (Maintenance of a computerized management system and access control to the recycling center, 2013). This also demands a change of approach from the top management to those dealing with operations. For the system to be integrated into the organization successfully, everyone in the organization should approach the whole maintenance management issues proactively and not reactively (Carroll & Wilmot, 2003). This avoids scenarios that arise when a problem that was imminent, but due to the reactive nature of the management, it was not identified. This may lead to a complete halt of the activities of an organization. The properly utilized maintenance programs perform preventive and predictive maintenance (Silverberg, & Taylor, 1999). This enables the plant to run uninterrupted, thus saving the organizations losses that could arise from the stall of operations.

This is achieved if the management is proactive and is able to use the potential of the maintenance technology.

In addition, maintenance systems are installed in an organization with a specific goal in mind. This varies from one organization to another due to various needs that may exist and that are unique to each organization. The existing systems have capabilities to solve most of these problems, but an organization must have the goal clearly spelt out to avoid purchasing an expensive system and then end up failing to use it appropriately. The goals outlined help the maintenance department determine what kind of information is expected from the system. It also helps outline what type of costs and historical data the system will be expected to track (Silverberg& Taylor, 1999). A clear goal also helps determine which equipment will be placed on the preventive maintenance program and how to track and order spare parts for the equipment. A goal also helps in the establishment of a benchmark that the system will use to rate the performance of the equipment and determine which ones are in need of repair and replacement (Birkland, 2006). The kind of information that the program requires in order to run successfully is immense. This requires proper prioritization to ensure that all the information that the system needs is put into the system (Silverberg & Taylor, 1999). This information includes all assets owned by an organization, employees, drawings, and vendor and manufacturing information, accounting data, preventive maintenance schedules, as well as other maintenance data that must be coded and entered into the system. All this data is important if the system has to play efficiently a role which it is designed for and was bought to play (Carroll & Wilmot, 2003).

Furthermore, due to the magnitude of this data, it is usually hard to enter all this data in one step requiring the process to be subdivided into a number of tasks that will take a certain period of time. Therefore, a priority must be set to ensure that the most important and basic information is first entered into the system (Silverberg, & Taylor, 1999). It is information that the system requires starting the operation and has an immediate impact in the organization. The completion of this initial stage will make the system run while the rest of the data will be included as time goes. Setting the right priorities in inputting data and following up the process as listed will ensure that when the system starts operating it will be stable and able to ruin the tasks it is programmed to play efficiently without failing because a failure may stop the whole process and the program may have started with the most critical processes in the organization (Carroll & Wilmot, 2003). These are the core operations of an organization whose collapse may halt every other operation in the entire organization. In addition, the management needs to use a think win-win approach (Silverberg, & Taylor, 1999). It is important due to conflicts that arise during the implementation with certain departments reluctant to delegate all their functions to the system. In such a scenario, a compromise is necessary to decide the best approach that will gradually incorporate all the functions of the department to the system without creating unnecessary rivalries within departments.

Consequently, the integration of the system receives the right support, and this makes it a seamless process. This demands a real creativity from the management or application of common sense to arrive at the right process that will gradually absorb all the functions of all departments into the system (Silverberg & Taylor, 1999). The opposition the system might receive from various people in the organization may require the management to apply listening skills so that the fears of those opposing the system are put into consideration. There are those people who perceive these systems as machines that cannot be trusted and whose main aim is to render people working in the organization redundant. These people need to be listened to, especially if explaining to them the benefits the system has does not work. Eventually, they slowly get to understand the position of the organization and the need to use technology to improve the efficiency (Allen, 1999). Most people resist change, and this poses one of the greatest problems that organizations have to deal with when introducing the computerized management systems. The management must listen to the source of the resistance that the subordinates have and then deal with the problem proactively with that in mind. Organizations that successfully integrate the system in their operations hold myriad open meetings within their workers and ensure that all fears that people have are addressed in an amicable way (Silverberg, & Taylor, 1999). Consequently, the workers cooperate and play their roles, thus ensuring that the integration process is smooth and that all the necessary data from their different department has been provided and properly entered into the system. Synergy has enabled many organizations to implement the system with minimal resistance (Raouf, Ali, & Duffuaa, 1993). This is because all the workers have felt part and parcel of the process because the organization has sought their help both intellectual and manual during the implementation process. This has made employees feel appreciated and view the system as one meant to make their work easier and increase their efficiency.

Consequently, after following the right procedure of setting goals and preparing employees for the imminent change, the management can identify the right management system and continue with the purchase and installation of the later. There are problems that face organizations that purchase a system that does not meet their needs (Molineaux, 1996). There are systems that offer different functions. Most organizations seek systems that are able to deal with the core functions of the organization. Core functions of most organizations are the integration between equipment records and store items to provide an up-to-date value of materials present, complete store room management functions for the single and multi-storage functions, and purchasing system module covering all necessary function of the purchasing department. Work order planning and scheduling functions including backlog, as well as preventive maintenance module and history of every equipment, are the main core functions a management system can be expected to play in an organization (Carroll & Wilmot, 2003). These are functions that prove hard for the management to track because they require a number of individuals working in different departments to compile their reports and hand them in to the management at certain intervals or when demanded.

Systems existing in the market offer different packages. Some may have all the features that cover the above features. Though, some may not integrate them properly if they are not used accordingly and customized to suit the needs of the organization (Sahoo & Liyanage, 2008). This arises from the inability of those responsible for entering data into the system and integrating it into the organization to exploit all the available features adequately. The management must decide on the best approach to use. Purchasing a system from an experienced and respected vendor in the market is always better than purchasing a system with very few users. This occurs because most of the companies that are in this business of making CMMS systems last less than three years in the business. Statistics show that only 30% of the companies making CMMS programs survive longer than that and only 10% produce systems with features that serve most of the core functions of the organizations, especially those with huge operations (Carroll & Wilmot, 2003).

Consequently, purchasing a system from a company that collapses soon is a huge loss to the buyer because the maintenance services will not be offered. There is also a high likelihood that the system will not be efficient, thus leading to a loss of funds and man hour spent entering the data into the system. This makes it important to purchase a system with many users and one that has been in operation for a couple of years. A huge base of users provides information about the effectiveness of a system, making it easier for new users to decide whether the system will fit their needs (Molineaux, 1996). The managements should draw a shortlist of systems from the trusted vendors and then decide which one should be purchased. When identifying a system, the priority should be given to such a system that can handle complex functions with those that can easily be handled manually considered later. The vendor should also provide proper orientation service to the employees. This implies that the management should also set aside money to be spent on educating the entire staff that will handle the system. Expanding capabilities, both in the hardware and software, implies that employees dealing with these systems should occasionally upgrade their skills to ensure that they handle the system changes that may be introduced when the vendor makes updates to the system (Autin, 1998).

The poorly trained workers may lead to a failure of the systems purchased and installed to undertake the duties they were to fulfill to perform properly. This has caused failures of a number of systems in many organizations because the workers do not have necessary skills to handle the new system. Necessary skills are required to ensure that integrating of the system in different departments is done properly and as dictated by the system. Due to the changes in the staff that take place in the organizations, it is necessary to install a system that is easy to operate and access (Singer, 1999). This helps to avoid problems that may arise if the employees that are trained to use it were replaced by the new workers with minimal skills in handling such systems. An ease to use system will pose few problems to the new staff and will take little time for them to understand it.

2.5    Installation and Implementation of CMMS

The implementation of a computerized maintenance management system takes place in phases. The first phase is the survey phase that consists of an interview between the management and a computer analyst to determine the right system for the organization. This involves checking the amount of work the new system is expected to handle (Singer, 1999). This stage exposes the problems the organization is experiencing from its current system of managing operations in the organization. Probable problems and constrains under which the new system to be installed is likely to experience are identified and evaluated to identify how well to deal with them. A proactive management also visits organizations that are already using the system that the organization has decided to buy (Computerized maintenance management system increases efficiency, 2011).  This visit helps them identify the effectiveness of the system and its limitations. The organization that is visited also helps solve questions, such as how many people are required to handle the system, limitations of the system in terms of the reports it generates, how the staff in the affected departments received the system, the length of the transition phase, what can be changed in the system once bought, how the organization feels about the system, and the quality of the sale service provided by the vendor of the system (Autin, 1998).The purchase of the system provides the first stage towards implementing it.

The second stage involves evaluating the benefits of the system and its cost. The right system should provide more benefits than the cost incurred to purchase and implement it (Singer, 2002). A budget containing the cost of software, the necessary hardware, and all other fees that will be incurred during the entire process of purchase and implementation should be included in the budget. A schedule showing the time frame should be drawn. The available hardware is studied, and the program modules are designed. The modules are designed to produce the required data. This is followed by testing the system to determine whether it works on the hardware and produces the results expected (Norman, 1997). Then the system is purchased if it passes the test it is subjected to.

The process of implementation provides the greatest challenge to any organization, due to the large volumes of data that has to be entered into the system (Vavra, 2005). This data must be compiled and then entered into the system. This process takes a long time, making it important to set priorities that will determine which data to compile and integrate into the system first. Most of the modules contained in the system require the data to be available in the system for them to function. These modules include a database that keeps track of all plant equipment, inspection and test data to add observations and measurement records to basic plant information, spare parts data in the form of electronic inventory, a single line diagram generation capability with drop down menus to facilitate elements and spare parts selection (Basta, 1996). A bar coding module that generates the barcode labels that can be attached to all the equipment in the plant should be included in the system (Applying bar code technology to today's maintenance systems, 1993). These labels identify each device and provide easy access to equipment information. Bar coding also eliminates entry of incorrect equipment information and identification (Eby & Bush, 1996).

The above procedure ensures that the right CMMS is bought and implemented. The compiled data should be entered into the system accurately to avoid having the system produce the wrong reports (Sahoo, 2008). This makes it very important to have all people in the implementation department in support of the new system so that they do not deliberately enter the wrong data. This has affected some organizations where the management has failed to acquire the support of the employees. Employees have deliberately entered the wrong data so that the system fails. The following act leads to the conclusion that the system is not effective as touted by the management (Hernandez, 2001, 06). It is also recommended that the new system should be learned in parallel to the old system that was in use, i.e. human maintenance, so that their effectiveness is compared. This helps to convince the workers that a computer system is more effective than human beings. The performance of the system should be evaluated after a given period to determine whether it meets the expectations of the management (Cooper, 1998). This allows the management to integrate more roles into the system once they are convinced that the system has performed the few roles assigned perfectly. One of the major problems affecting most organizations that implement a computerized management system is a failure to utilize all the features of the system (Cleaveland, 2005). This is a loss to the organization because it paid for all the features including the ones it does not use. The cause of this is workers who are not well trained in utilizing the system (Eby & Bush, 1996).

## 2.6    Evaluation and Design of CMMS

The design of a computerized management system should be done with the end in mind. This implies that before designing a system, a number of factors ought to be considered. The first factor is the market needs (Raouf, Ali, &Duffuaa, 1993). This is based on the organizations that are targeted as potential buyers of the system. If the system is designed for a certain organization the needs of the organization must be considered before the system is designed. The prevailing competition in the global market where the targeted organizations operate should be considered (Tennessee inventors develop computerized maintenance management system, 2008). This is to integrate the current and future needs of the organization so that the system can accommodate the likely changes that occur when the organizations try to upgrade their operations so that they remain ahead of their rivals (Polakoff & Laughlin, 1992). A failure to consider this aspect leads to the development of a system that becomes obsolete within a short period of time. The ability of the information system department to implement the system by supplying the necessary data is also an important factor to consider. It is more important if a system is designed on the request of a certain organization. The vendor must consider the readiness and the ability of the department so that the developed design fits seamlessly (Klusman, 1995). The hardware and other supporting infrastructure available must be considered. This should be done to avoid developing a system that will require a complete overhaul of the available hardware that might be very costly to an organization. Finally, the ability of organizations targeted to install, implement, and utilize the CMMS should be considered (Autin, 1998). This is important because not all industries developed them in the same way. Organizations that deal with information technology are likely to install and utilize this system easily because their employees have the computer knowhow while other industries might struggle due to little knowledge of computers that their staff has, which makes the implementation and utilization process cumbersome (Elliott, 2000).

The management of the mainframe, management of the distributed environment or client server, management of the desktop environment and management of the network are key areas of a CMMS that must be well designed to ensure that organizations that use the system are able to manage them properly (Carroll & Wilmot, 2003). These are the key areas that determine the effectiveness of a CMMS in an organization. They are responsible for the collection of data, as well as storage and processing of reports. The design of the system must ensure that those using the system will be able to manage the network, the mainframe and the desktop environment easily (Antonacci, 1994)). This is to avoid issues that arise due to inability of the staff handling the system to obtain information from the system due to its complex nature. This poses a challenge to a system developer to ensure that the system is advanced in its features and functions, is compatible with the available hardware and is user friendly (Emmott, 1999). Some vendors have developed very complex systems with many complex features that have flopped in the market because they are not user friendly. Organizations desist from purchasing systems that will pose challenges to their staff in terms of utilizing and maintaining (Autin, 1998).

2.7     Benefits of Using CMMS

The objective of maintenance is to reduce the costs incurred when operations slow down or stop due to problems with the equipment used in the production process. To avoid this, all organizations have people responsible for maintaining the equipment used by an organization. This stretches further where an organization owns vehicles that it would want to monitor to avoid incurring losses that arise from engine breakdowns and misuse of the vehicles by the drivers (Leavitt, 2007). For many years, this has been done by human beings; though, the effectiveness has been low. This has been solved by the use of the computerized systems that oversee the entire operations of an organization (Mullin, 1992). Maintenance management is a continuous process that runs parallel with the production process. This is because problems arise during the production process that require maintenance to rectify and prevent them escalating to the extent of halting the entire production process. These systems have the ability to produce reports showing how the equipment has been performing at any given period (Koss, 1992).

        The quality of the products produced must be tracked to avoid the production of low quality goods that breach the set quality standards in the market. This implies that samples must be collected occasionally and tested to determine their quality. Management systems can be

developed to undertake this role and release reports of the quality produced throughout the day (Stoller, 2006). They have the ability to test the quality of all products released and where a product of low quality is detected, it is ejected from the system or a warning is sent to the quality department so that the defective product can be removed from the rest of the products (Hammer, et al, 1992). This is more effective than what was done earlier when people had to collect samples and test their quality. This is because defective products could still pass through without the detection and land in the market, which would sometimes cost organizations huge losses if consumers sued them due to the production of poor quality goods (Andel, 1996, 09). Computerized maintenance management systems have saved organizations huge losses and kept their quality standards high boosting their image in the market (Hammer, et al, 1992).

Asset management has often posed problems to management of huge organizations that have many branches all over the world. This has led underutilization of the available assets because some are not accounted for. Having records of all assets available to an organization has proved important because it is possible to determine how to use all existing assets to achieve growth objectives of the organization (Fulkerson, 2007, Dec). Systems also prevent theft of these assets by employees who take advantage of the inability by the management to track all assets available. This has increased productivity and reduced the loss of the valuable assets owned by organizations (White, 2004).

Maintenance work has become easier and highly effective because computer systems track performance of equipment and report drops in productivity. These drops indicate a need to check the equipment for problems before its performance deteriorate (Duell & Beck, 2003). The system also keeps records of the available spare parts and ensures that those that are running out of stock are ordered in time. This eliminates instances where a machine breaks down but cannot be repaired due to the lack of spare parts in the store. This has saved many organization losses arising from equipment downtime (Slaichová & Marsíková, 2013). The system stores information about maintenance routine of equipment as recommended by the manufacturer. The system reminds the maintenance department about the dates to ensure that the equipment is serviced in time to avoid breakdowns.

Computer systems produce reports of all operations that are integrated into it. These reports provide the management with the valuable information that is used when they set targets for the various departments. These reports are also used to determine which departments are underperforming and which need to be changed to improve their productivity (Maintenance management software computerized maintenance (cmms), 2013). The system also helps in scheduling of duties. The tasks that it schedules include the allocation of manpower, management of fluctuations in the workload, scheduling of work, management of the manpower pool, control of backlogs, and monitoring flow of work orders (Maintenance system improves manufacturing performance, 1996). Scheduling enables the management to achieve high productivity from its employees. The increased productivity leads to high profits and high efficiency in utilization of the available resources. The system is able to set the sequence of tasks creating a program that ensures that all tasks are catered for (Westerkamp, 2006). This has reduced wastage of manpower in the organization, which has enabled the management to reduce unnecessary overheads by working with the right number of employees. This could not have been possible if it were not used for the abilities of the management systems.

## 2.8    Related Works

Several works in literature are used the base for this research. One of the most related works is published by Kevin Kaiser and Nagi Gebraeel (2009). The authors proposed Sensory Updated Degradation-based Maintenance (SUDM) policy to perform predictive maintenance. SUDM collects degradation signals from the equipment such as sound, vibration or temperature. Based on these signals the Residual Life Distribution (RLD) is updated. RLD is a mathematical function which describes probability of equipment life time. The remaining life time of the equipment is estimated by calculating the expectation of RLD. In (Ming-Yi et al. 2010), the authors tried to improve SUDM by proposing Statistically Planned and Individually Improved Predictive Maintenance (SPII PdM). Their model is composed of two phases. In first phase, they globally compute statistical parameters for all equipment's. Afterward, every equipment will have its statistical parameters updated individually. The most recent work (Hong et al. 2014) focused on the local characteristics of maintenance. The author argued that statistical parameters of equipment maintenance depend on degradation of each component in multi-components equipment. Also, they distinguished between two types of equipment failure risk management. For some equipment, maintenance should be delayed until failure time is very close which is called Risk-Seeking management. The other is called Risk-Averse management where valuable equipment should be maintained as soon as possible.

Chapter 3:     Methodology

In this chapter, we describe the main methodology used in this research. The primary features of the adopted methodology are highlighted with emphasis on its functionality and performance. How the adopted methodology is implemented to solve the research problem is described in the next chapter.

## 3.1     Introduction

Artificial Neural Networks (ANN) is one of the most powerful techniques in artificial intelligence field. It is inspired by how human brain operates. This technique was created at the end of forties of the last century. McCulloch and Pitts (1943) were first to show how this technique can be very powerful to solve many difficult problems. ANN is a network of interconnected small elements called neurons. Each one of these elements can perform specific computation that lead to the targeted goal. The input data is provided to these elements and the desired output should be expected. Nowadays, Artificial Neural Networks are used in many fields such as medicine, engineering, business management and education.

## 3.2     ANN Elements

As mentioned above, artificial neural networks are composed of neurons. These neurons take a list of input data and they used these data to produce an output. One of the first models of neurons is called Perceptron. In this model, the input data is multiplied by a list of weights and the result of this multiplication would be accumulated for output computation. If the sum of weighted inputs is larger than some threshold, the neuron output would be equal to one. Otherwise, it would be zero.

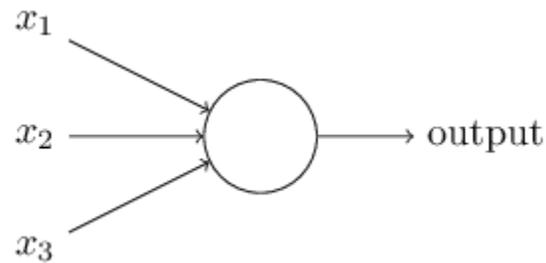

Figure 3.1: Diagram of one neuron.

$$\text{output} = \begin{cases} 1, & \displaystyle\sum_{i}^{3} w_i x_i > \text{threshold} \\ 0, & \displaystyle\sum_{i}^{3} w_i x_i < \text{threshold} \end{cases}$$

The above figure describes the basic mathematical model for perceptron. In this model, the neuron output can be modified by changing the threshold value or weight values assuming that the input data is the same. This process of changing values is called Learning.

The following example is provided to show how the perceptron can be taught to produce the desired output. Imagine, $x_1$ represents the equipment Age and it can take two values, New or Old. The New value is denoted by zero; while the old value is denoted by one. Also, let $x_2$ presents the equipment Sound which can take two values, Normal or Not Normal. The Normal value is denoted by zero; while the Not Normal value is denoted by one. Likewise, $x_3$ represents the equipment temperature which can take two values, Cold or Hot. The Cold value is denoted by zero while the Hot value is denoted by one. Let the Output be the decision of performing the maintenance. If perceptron output is equal to one, then the maintenance should be performed on the equipment.

Now, imagine that we have equipment which is old ($x_1 = 1$), with not normal sound ($x_2 = 1$) and it is hot ($x_3 = 1$). Clearly, in this situation we would like to perform the maintenance (output = 1) before the equipment get damaged. In another situation, we have equipment which is old ($x_1 = 1$), with normal sound ($x_2 = 0$) and it is cold ($x_3 = 0$). In second situation, we don't need to perform maintenance (output = 0) since there is no indication of equipment failure. Therefore, values of $w_1$, $w_2$, $w_3$ and threshold should be chosen in a way that lead to output = 1 in the first situation and output = 0 in the second situation. A valid assignment of values can be $w_1 = 1.5$, $w_2 = 0.85$, $w_3 = 2$ and threshold = 2.7. In first situation:

$$\sum_i^3 w_i x_i = w_1 x_1 + w_2 x_2 + w_3 x_3 = (1.5 \times 1) + (0.85 \times 1) + (2 \times 1) = 4.35$$

Which is larger than 2.7. As a result, output = 1. In the second situation:

$$\sum_i^3 w_i x_i = w_1 x_1 + w_2 x_2 + w_3 x_3 = (1.5 \times 1) + (0.85 \times 0) + (2 \times 0) = 1.5$$

Which is less than 2.7. Therefore, the perceptron output = 0. Values of weight should be dependent on the equipment type. In previous the assignment, the most important input is $x_3$ because it has the largest weight. For other type of equipment, maybe $x_2$ should be the most important input. Therefore, any Learning process should take external parameters into consideration to achieve the best weight and threshold values assignment.

3.3    Sigmoid Neuron

The mentioned perceptron model for neurons can be very useful in many applications. However, it suffers from two important limitations. The first limitation is the type of input. Perceptron assumes inputs to be binary ($x$ has only two values which are 0 and 1). However, we may face a situation where input has to have more than two values. For instance, in the previous example, the Age input ($x_1$) can be represented by the number of hours the equipment has been used. The same can be applied to the other inputs.

The second limitation relates to how the output is calculated. Using perceptron calculation leads to binary output as well. It will be very useful if the output can have multiple values between 0 and one. One way to achieve this is by introducing more threshold values:

$$\text{output} = \begin{cases} 1 \, , & \sum_i^3 w_i x_i > \text{threshold 1} \\ 0.7 \, , & \sum_i^3 w_i x_i > \text{threshold 2} \\ 0.4 \, , & \sum_i^3 w_i x_i > \text{threshold 3} \\ 0 \, , & \sum_i^3 w_i x_i < \text{threshold 3} \end{cases}$$

It is clear that such approach has limited application and introduces the challenge of learning the best set of threshold.

Sigmoid neurons can solve these limitations by using the concept of Activation Function (*f(x)*). The output of these neurons is calculated by applying the activation function to the sum of weighted inputs and threshold:

$$\text{output} = f\left(\sum_{i}^{3} w_i x_i + b\right)$$

where *b* stands for the bias (threshold). The activation function in these neurons is called the Sigmoid function (hence the name):

$$f(z) = \frac{1}{1 + e^{-z}}$$

Which has the following shape:

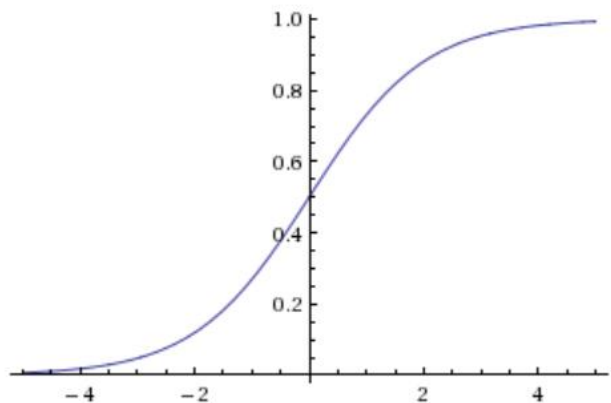

Figure 3.2: Sigmoid function output.

It is obvious that the output of sigmoid neuron can have any value between zero and one depending on the inputs, their weights and threshold assuming that $z = \sum_{i}^{3} w_i x_i + b$.

3.4    Hyperbolic Tangent Neuron

Some situations may require the output of neurons to have negative values. At the same time, the behavior of Sigmoid function is very preferable. We can extend the sigmoid neurons to generate output with values between -1 and 1 by using the following activation function:

$$f(z) = \frac{e^z - e^{-z}}{e^z + e^{-z}}$$

Which has the following shape:

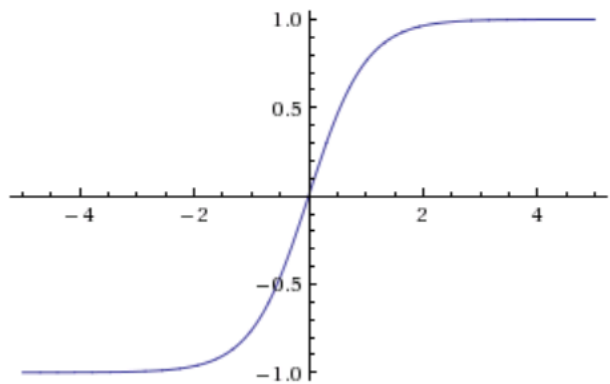

Figure 3.3: Hyperbolic Tangent function output.

Allowing neurons outputs to have negative values may improve the learning process. But, it is not a necessary condition. Sigmoid neurons have shown very powerful performance in many general applications; while Hyperbolic Tangent neurons usually improved the performance in specific situations. Both of these neurons will be investigated in this research.

3.5     Neural Network Architecture

As the name suggests, ANN is a network of connected neurons. Usually, Sigmoid neurons are used. In this network, some neurons will have their input coming from external data source (i.e. Condition-Based Monitoring system). Others will take the output of other neurons as their input. Repeating this way of connecting neuron will lead to multiple layers neural networks. Neurons in the first layer use external data as their input to calculate their output. Neurons in the second layer use the outputs of first layer neurons as their input. This process keeps repeating until neurons in the last layer calculate their outputs. The last layer outputs are the overall neural network output.

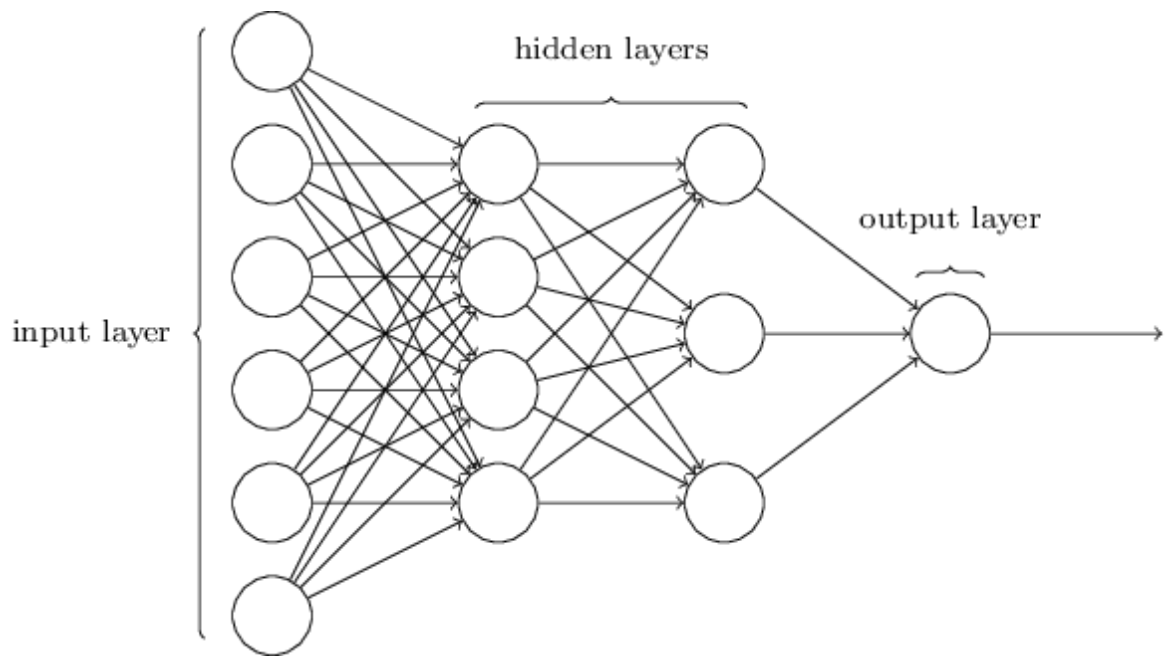

Figure 3.4: Multiple layers neural network.

The first layer is called the input layer and the last layer is called output layer. Layers between input layer and output layer are called hidden layers.

It is evident that neural network has much powerful capabilities in generating any type of decision more than single perceptron. Mathematically speaking, neural networks are called Universal Approximate. It means that neural network have the ability to approximate any mathematical function in the whole universe. From philosophical point of view, we can assume that neural networks can perform any mental or cognition computation which a human brain can perform if the neural network is large enough and complex enough.

3.6     Neural Network Learning

In section 3.2, we saw an example of teaching a perceptron how to generate a binary maintenance decision based on the binary input of equipment parameters. In the continuous form of neurons (Sigmoid or Hyperbolic Tangent), we can use their continuous feature to have an automatic learning mechanism. In the new learning model, neurons will start with random values of weights and bias (threshold). Then, automatically they keep update these values based on their run time experience.

*Error Function*

Every neuron in the neural networks updates their weights and bias based on estimation error. For example, imagine a boy trying to throw a ball in a basket. The neural network inside this boy's brain keeps modifying its weight every time the boy misses the basket by values depending on how much the boy missed the basket. This process keeps happening until the boy scores. The same process is applied in artificial neural networks.

Having said that, an exact error function should be defined so that weights updates can be calculated. By having the neural network output and the desired value, several error functions can be formulated. One of the most used functions in literature is Squared Sum Error:

$$E = \frac{1}{2} \sum_{t}^{T} (d_t - o_t)^2$$

Where $o_t$ is the neural network output at time $t$ and $d_t$ is the desired output (what the neural network should estimate) at time $t$. Any learning algorithm should try to minimize this error.

*Gradient*

Minimizing any continuous and differential function depends on a mathematical concept called Gradient. Since Sigmoid (or Hyperbolic Tangent) neurons are used, the error function in previous section is both continuous and differential. Therefore, Gradient can be employed to minimize the error and calculate the best update for weight values. One way to describe the gradient is to imagine it as a vector which point to the direction of the maximum value of the function. The negative gradient points to the minimum value of the function. The gradient of error function is calculated by taking the partial derivate of error function with respect to each weight in the neural network:

$$\nabla E = \begin{bmatrix} \dfrac{\partial}{\partial w_1} E \\ \dfrac{\partial}{\partial w_2} E \\ \dfrac{\partial}{\partial w_3} E \\ \vdots \\ \dfrac{\partial}{\partial w_N} E \end{bmatrix}$$

The weights in this formula represent weights of all neuron in the whole network. It is very difficult to calculate the gradient using the classical mathematical approach.

*Back-Propagation Learning*

Because of the complexity of error function in large neural network, computing the derivative based on single weight is very tedious. Therefore, more practical method should be used. The next flowchart (Figure 3.5) shows the general steps of back-propagation learning.

Each step in learning is called Epoch. The first step is to initialize the neural network with required number of neurons and layers structure. Then, weights and bias values of each neuron are chosen randomly. After that, the output of the neural network is calculated based on the input data. First, the input is fed to the neuron in the input layer. Then, the output of input layer neurons is fed to the second layer neurons. This process keeps repeating until neural network output is calculated at the last layer neuron.

By having the neural network output, we can calculate the error value based of sum squared error function. At this point, we need to check if the error is below sum threshold or if the maximum allowed number of epoch is reached. If any of these conditions is correct, the learning process is stopped. If these conditions are not satisfied, weights of neurons are updated using the learning rule.

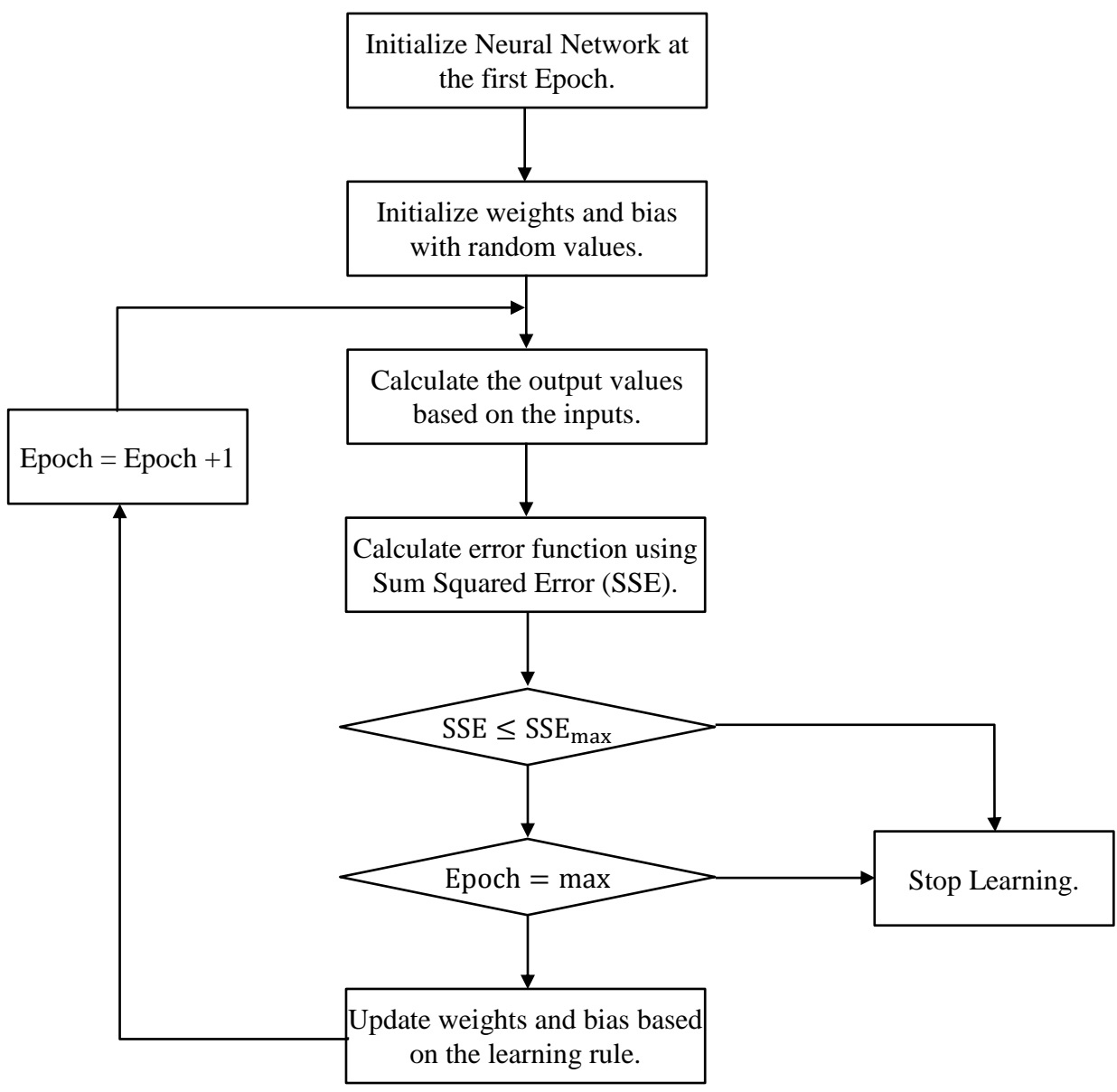

Figure 3.5: Backpropogation learning for neural networks.

The learning rule depends on idea of propagating the error backwards from the neurons in the output layer to the neurons of preceding layers. The new weights of each neuron are calculated as follow:

$$w_i(t + 1) = w_i(t) + \gamma(1 - o_i)o_i err$$

The weight at time t+1 is equal to the weight at time t in addition to the update value. The update value is equal to the learning rate ($\gamma$) multiplied by $(1 - o_i)o_i$ where $o_i$ is the neuron output. Lastly, the error coming from front layer neurons ($err$) is multiplied by the update value.

Chapter 4:    Neural Condition-Based Maintenance (N-CBM)

This chapter discusses how artificial neural networks technique is used to perform condition-based maintenance. At the beginning, modeling of component degradation will be presented. After that, maintenance cost calculations will be deliberated. Later, the classical approach of estimating inspection interval will be discussed. Lastly, the adopted neural network implementation will be delivered.

## 4.1    Component Stochastic Degradation

Most works in literature use stochastic approach to model component degradation. The amount of degradation experienced by any component can be considered as a function of time $X(t)$. There are many uncertain factors which may affect the value of $X(t)$. Therefore, stochastic approach is used to calculate $X(t)$. The most adopted stochastic process for this sort of modeling is Gamma process where the probability density function is calculated as follow:

$$p(x(t)|at, b) = b^{at}[x(t)]^{at-1} \frac{e^{-bx(t)}}{\Gamma(at)}$$

Here, $a$ and $b$ are gamma distribution parameters. Function $\Gamma()$ is gamma function. Also, the mean of this process is $at/b$ while the variance is $at/b^2$.

Having this degradation process in mind, we would like to define inspection time $(T_I)$ where the inspection is performed at the end of this interval. Keep in mind that as time progress, component degradation increases. Also, whenever a component replacement is performed, the degradation process will restart from zero as seen in the following figure:

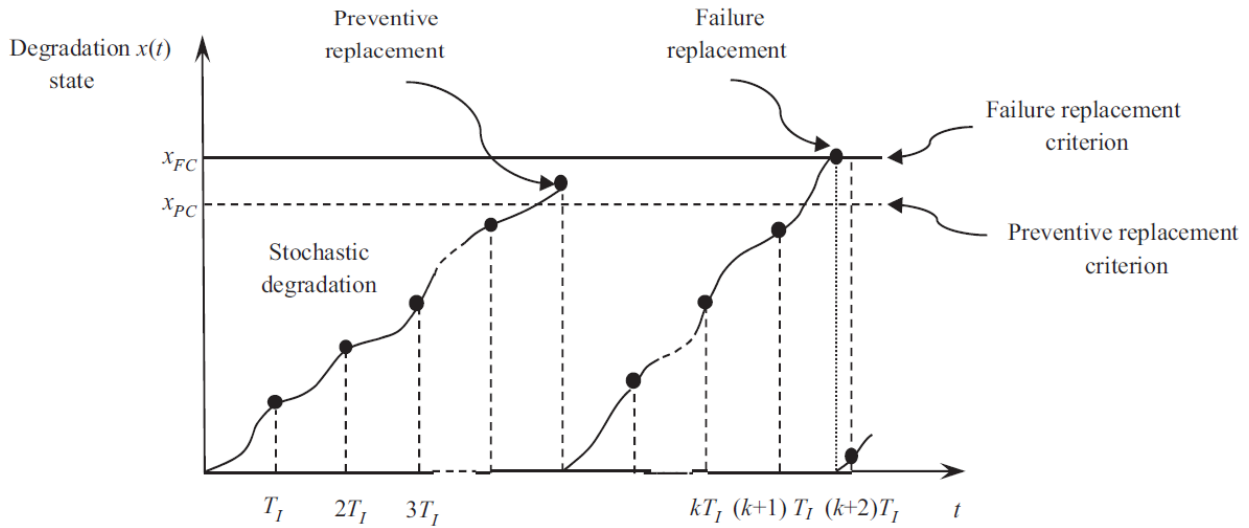

Figure 4.1: Degradation process over time (Hong et al, 2014).

Two types of maintenance can be defined. First type is Preventive Maintenance where the component replacement is performed before its failure. Second type is Failure Maintenance where the component is replaced only when it broke. Two values are associated with these types of maintenance. These values represent the amount of degradation which requires the maintenance. If the degradation is larger than or equal $x_{PC}$, the preventive maintenance can be performed; while failure maintenance is performed when degradation reaches $x_{FC}$.

## 4.2    Maintenance Cost

There are three types of activities which generate cost. The first activity is inspection. Any component should be inspected from time to time to check its degradation state. This activity generates inspection cost ($C_I$). The second activity is preventive replacement where the component is replaced before it fails. This activity generates preventive cost ($C_P$). The third activity is failure replacement where the component is replaced after its failure. This activity generates failure cost ($C_F$).

In general, failure cost is most expensive with multiple orders of magnitude compared to other types of cost. Preventive cost comes second with higher value than inspection cost. Keep in mind that, inspection cost in generated whenever inspection is performed. And, the inspection is performed every $T_I$ until an inspection leads to preventive replacement where preventive cost is added as well. Therefore, the total cost over operation time ($T$) can be calculated as follow:

$$C_R(T_I) = \frac{\left(\sum_i^n\left[C_I e^{-\gamma i T_I}\right] + \sum_i^{n_P}\left[C_P e^{-\gamma t_{Pi}}\right] + \sum_i^{n_F}\left[C_F e^{-\gamma t_{Fi}}\right]\right)}{T}$$

As seen in the previous equation, the total cost is combined of three mentioned types of costs. The discount rate ($\gamma$) is introduced to add the effect cost reduction over time. The total number of performed inspections is denoted by $n$; while $n_P$ represents the total number of preventative replacement and $n_F$ represents the total number of failure replacement.

## 4.3    Classical Condition-Based Maintenance

The total cost formula emphasis that value of $T_I$ plays crucial role in minimizing the total cost. Therefore, any good maintenance manager will try to find the optimal value of $T_I$ which results in the minimum total cost on average. To find this optimal value, simulation approach is used in literature. The main idea is to try every possible value of $T_I$ in simulation environment many times and to adopt the best value on average. The following flow chart describes simulation.

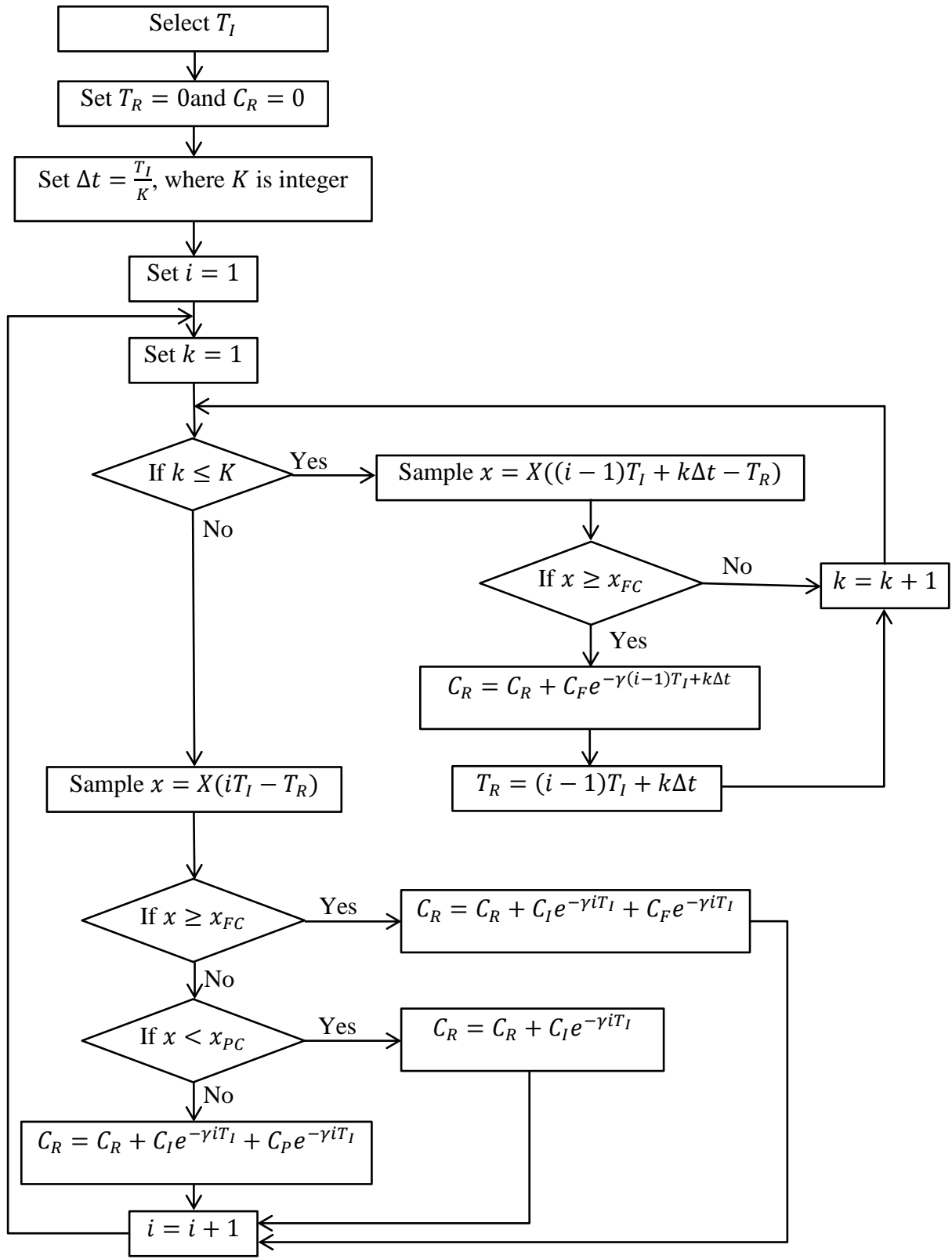

Figure 4.2: 4.3 Classical condition-based maintenance flow chart.

At the beginning of simulation, the minimum possible value of $T_I$ is selected. The other degradation process parameters are fixed for all simulation experiments. These parameters are operation time, inspection cost, preventive cost, failure cost, preventive degradation level, failure degradation level, discount rate and gamma distribution parameters (a and b). A simulation will be conducted based on the selected $T_I$.

The first step in this simulation is to initialize both of $T_R$ and $C_R$ to zero. The variable $T_R$ represents the last time the component was replaced. Second step is to define $\Delta t$ time interval which will be used during the simulation to check if any component failure has happened. Keep in mind that this check does not happen in reality. It is only for simulation purpose. In reality, the component just fail and the management will notice its failure. The integer $K$ is defined in advanced by the designer.

At this point, $i$ and $k$ is initialized to one. If $k$ is larger than $K$, the simulation will proceeds directly to perform inspection at $T_I$. Otherwise, the degradation state will be sampled at $X((i-1)T_I + k\Delta t - T_R)$. If the degradation state is larger than or equal to $x_{FC}$, a failure cost will be added to the total cost and new value for $T_R$ will be assigned as $(i-1)T_I + k\Delta t$ to register the last time the component was replaced. The last step will keep repeating until $k$ is larger than $K$.

Now, the actual inspection will be performed. Therefore, the inspection cost will be added to the total cost regardless of degradation state. There are three region of degradation state. Either it is larger than or equal to $x_{FC}$ which means that the failure cost will be added to the total cost as well; or the degradation value is less than $x_{PC}$ which means no extra cost will be added to total cost. The third region is covering values between $x_{PC}$ and $x_{FC}$. If the degradation state in this region, preventative cost will be added to the total cost.

4.4     Neural Condition-Based Maintenance

This research proposes to use machine learning technique which is very effective in many fields. This technique is artificial neural networks which an attempt to imitate the ability of human brain reasoning. Imagine that you have the ability to assign an employee to every machine for maintenance purpose. This employee uses his advanced reasoning abilities to estimate the degradation level of the machine. Based on his assessment, the employee will be able to assign the best inspection interval. This approach is very costly and unrealistic.

Artificial neural networks provide us with human brain abilities without the extra cost. Also, it can be tuned to be focused on one type of application and to be resized in any way necessary. In this research, an inference engine based on artificial neural network is developed to estimate the degradation level of the specific component. This engine takes the current degradation level as an input to produce the maintenance decision. This decision is composed of two elements. The first element is the expected degradation level in near future. The second element is the estimated time to the failure point.

Similar to human brain, artificial neural networks requires training and learning before it can be efficiently used. In the classical maintenance management approach, the simulation phase is used to find the optimal inspection interval. Here, training phase is used to teach the neural network how to accurately estimate the degradation level of the component under consideration. Also, this phase provides us the best risk factor of our approach which is the estimation error during the training. This risk measure can be used to generate the optimal maintenance decision which results in much lower cost. Also, this risk measure can be used for any required analysis for the overall risk management of maintenance operations.

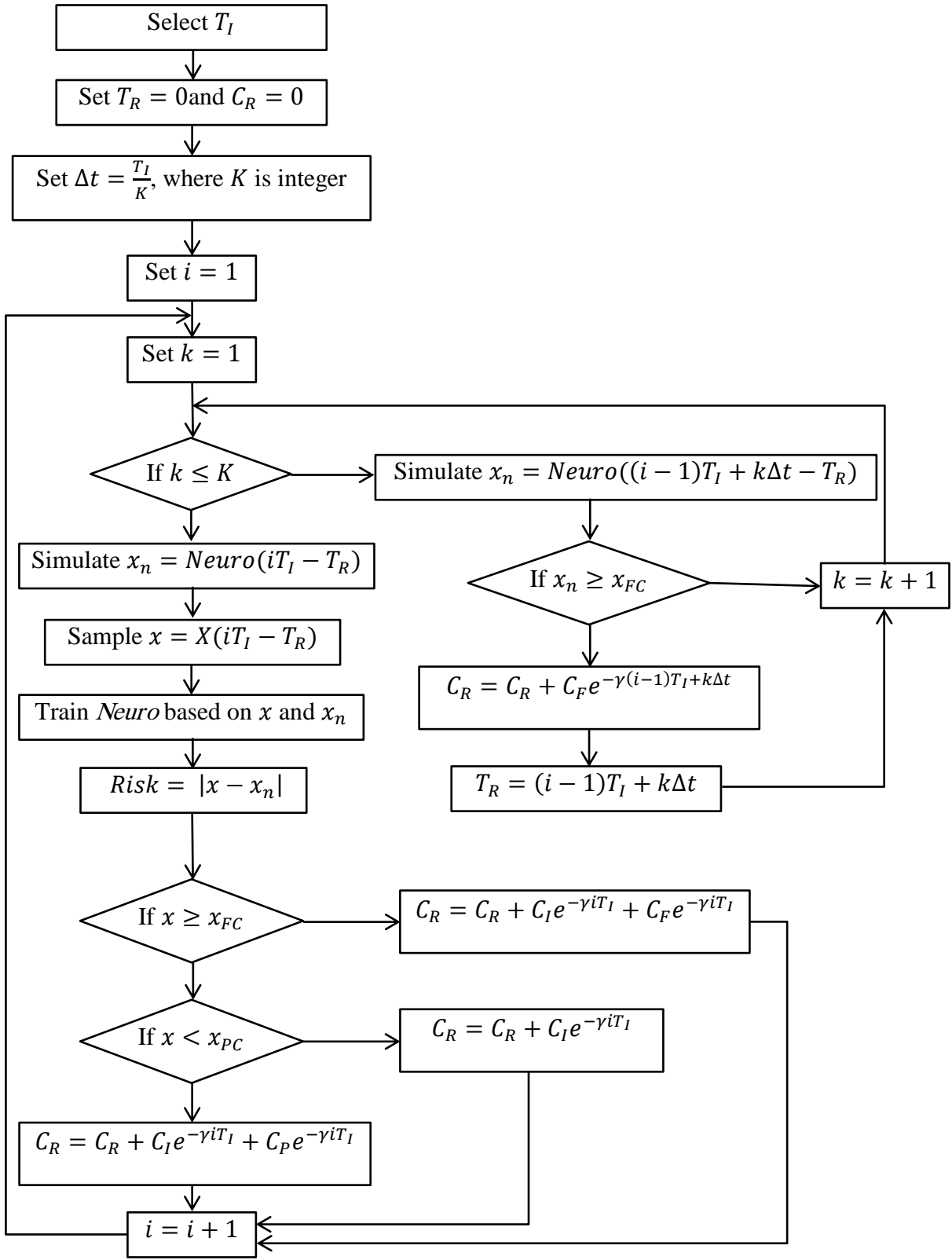

Figure 4.3: Neural condition-based maintenance flow chart.

As said before, neural training is used during the simulation to teach the proposed engine how to accurately estimate the degradation level. The same simulation process used in classical maintenance approach can be used in our approach with some modification as depicted in figure 4.3. At the beginning of training, the minimum possible value of $T_I$ is selected. The other degradation process parameters are fixed for all training epochs. These parameters are operation time, inspection cost, preventive cost, failure cost, preventive degradation level, failure degradation level, discount rate and gamma distribution parameters (a and b). A training will be conducted based on the selected $T_I$.

Similar to the previous approach, the first step in training is to initialize both of $T_R$ and $C_R$ to zero. The variable $T_R$ represents the last time the component was replaced. Second step is to define $\Delta t$ time interval which was used during the previous approach simulation to check if any component failure has happened. In our case, this variable is used to estimate the degradation level. Here, $i$ and $k$ is initialized to one. If $k$ is larger than $K$, the simulation will proceeds directly to perform inspection at $T_I$. Otherwise, the degradation state will be simulated based on $Neuro((i-1)T_I + k\Delta t - T_R)$ which is the proposed inference engine. If the estimated degradation state is larger than or equal to $x_{FC}$, a failure cost will be added to the total cost and new value for $T_R$ will be assigned as $(i-1)T_I + k\Delta t$ to register the last time the component was replaced. The last step will keep repeating until $k$ is larger than $K$.

When $T_I$ has elapsed, an inspection will be performed ($x$). Also, degradation state will be simulated ($x_n$). The risk measure can be calculated by finding the difference between the actual degradation state and the estimated degradation state. We would like to reduce the risk as much as possible. Hence, it is essential to perform training epoch based the observed risk and actual degradation state. The remaining of cost calculation is similar to classical condition-based maintenance. The inspection cost will be added to the total cost regardless of degradation state. There are three region of degradation state. Either it is larger than or equal to $x_{FC}$ which means that the failure cost will be added to the total cost as well; or the degradation value is less than $x_{PC}$ which means no extra cost will be added to total cost. The third region is covering values between $x_{PC}$ and $x_{FC}$. If the degradation state in this region, preventative cost will be added to the total cost.

Chapter 5:     Results and Discussion

In this chapter, extensive evaluation of the proposed condition-based maintenance approach will be presented. To highlight the strength of the proposed approach, direct comparison with one of the most recent work (Hong et al, 2014) will conducted. Deep discussion for each experiment is provided.

## 5.1     Experiment Implementation

For comparison reason, the same experiment settings used in (Hong et al, 2014) is adopted in this thesis. To simplify the discussion of experiment implementation, an overall run from the main simulation file is presented as step-by-step process. The source code for the remaining function files is presented in the appendix. The main experiment steps are:

*Clearing Memory and Workspace*

This is the first step in any scientific simulation to prevent mixing data results from different experiments.

```
clear all;
clc;
```

*Setting Number of Simulation Experiment*

As mentioned before, the simulation experiments have been conducted for several times. This is done to show the strength of the proposed solution on average.

```
N = 5000;
```

*Setting Gamma Parameter*

This parameter controls the effect of inflation in cost calculation. If Gamma is zero, this means that the cost at the end of the year is the same as the cost at the beginning of the year. However, this is not the case in the reality. Inflation reduces the cost as time progress. Here, Gamma represents the rate of cost reduction. Two set of simulation experiments were conducted. One set is conducted where Gamma is zero and another is conducted where Gamma is 0.05.

```
gamma = 0;
```

*Setting Simulation Parameters*

These are the parameters used by Hong et al (2014). In their paper, the authors simulated piping system in nuclear reactors. The degradation process is due corrosion resulted from heating which affect the wall thickness of the pipes. At installation, pipe thickness is 6.5 mm. When thickness reaches 4.5, preventive maintenance is performed. If the thickness is less than 3.41, failure maintenance is performed. The first parameter is simulation time (T) which is in years.

```
T = 50;
```

All cost parameters are normalized. Inspection cost:

```
C_I = 0.1;
```

Preventive cost:

```
C_P = 1;
```

Failure cost:

```
C_F = 10;
```

Preventive maintenance threshold:

```
x_PC = 2;
```

Failure maintenance threshold:

```
x_FC = 3.09;
```

Gamma distribution parameters:

```
a = 10 / 9;
b = 100 / 9;
```

*Collecting Neural Training Data*

To train neural network, we need to generate training data. Here, training data is generated by running stochastic simulation experiments and collecting all aspects of maintenance operations.

```
data = get_nn_train_data(T,x_PC,x_FC,a,b);
```

*Train Neural Network*

This is the most important step in our proposed condition-based maintenance. Here, the neural engine is trained to be able to accurately estimate the degradation level of components.

```
[net, err, inputs, targets, outputs, errors, trainTargets, valTargets, testTargets, tr] =
create_nn(10, data);
```

Calculate and present Training, Validation and Test performance of the training process.

```
performance = perform(net,targets,outputs)
trainPerformance = perform(net,trainTargets,outputs)
valPerformance = perform(net,valTargets,outputs)
testPerformance = perform(net,testTargets,outputs)
```

View the Neural Network architecture.

```
view(net)
```

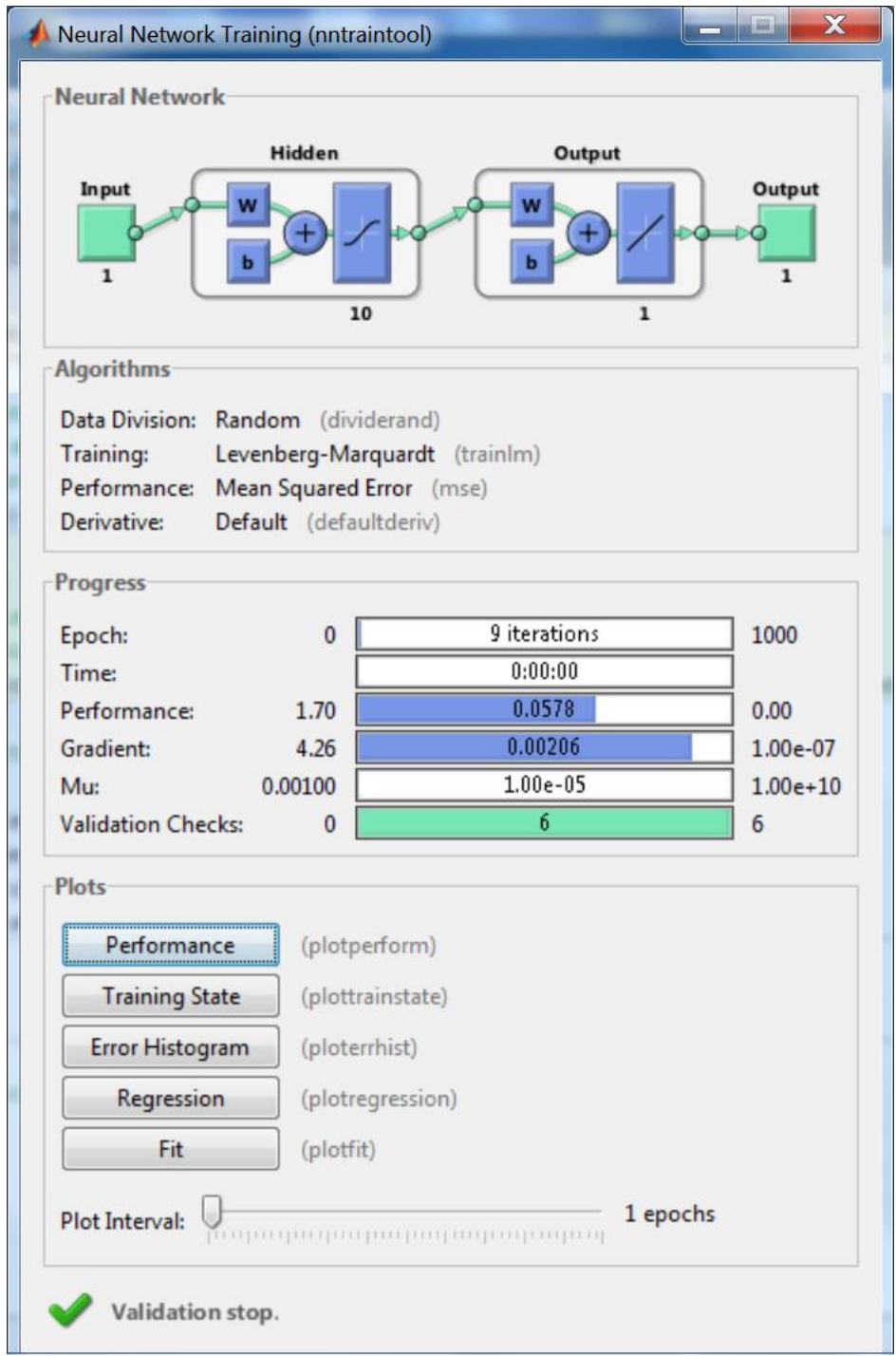

Figure 5.1: Neural Network architecture and learning process.

Next, we plot several Neural Network performance figures which show the several overall performance aspects graphically.

```
figure, plotperform(tr)
figure, plottrainstate(tr)
figure, plotfit(net,inputs,targets)
figure, plotregression(targets,outputs)
figure, ploterrhist(errors)
```

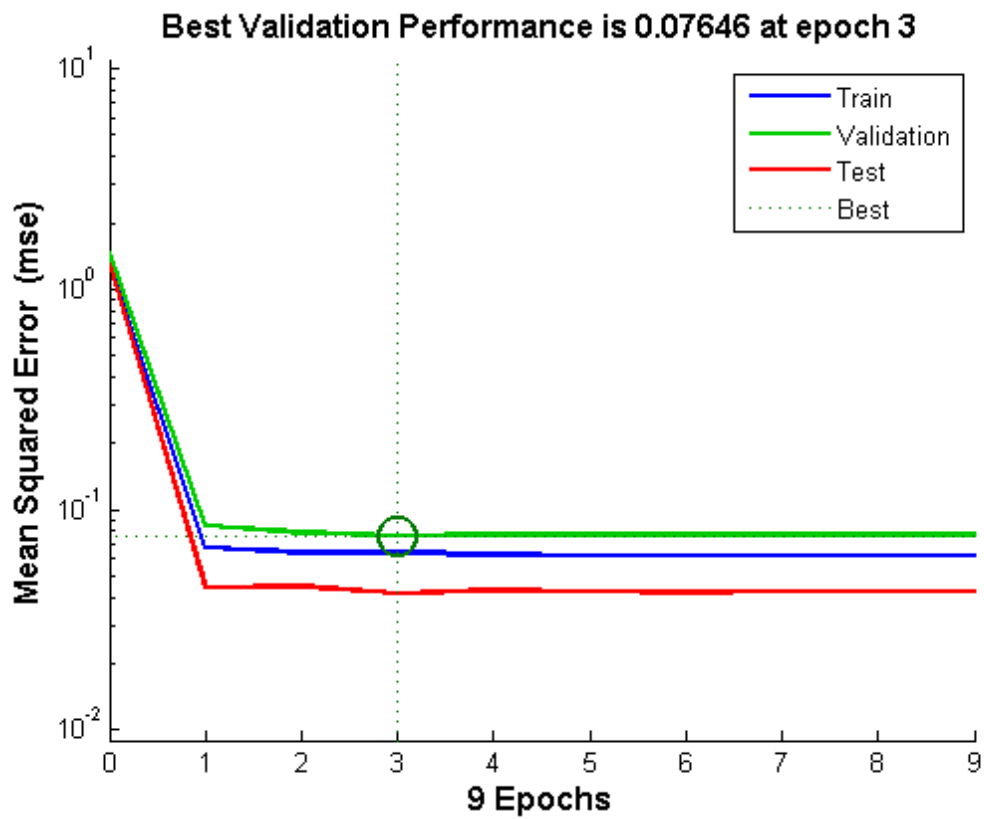

Figure 5.2: Mean squared error during the learning process.

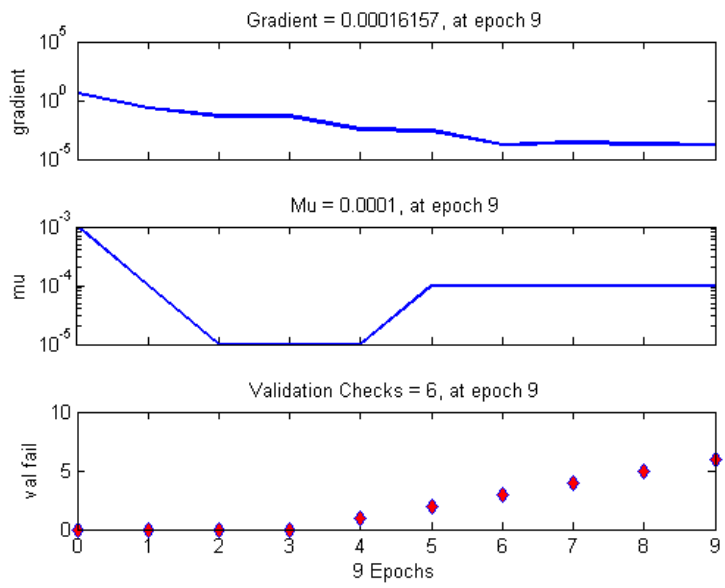

Figure 5.3: The gradient evolution as learning process progress.

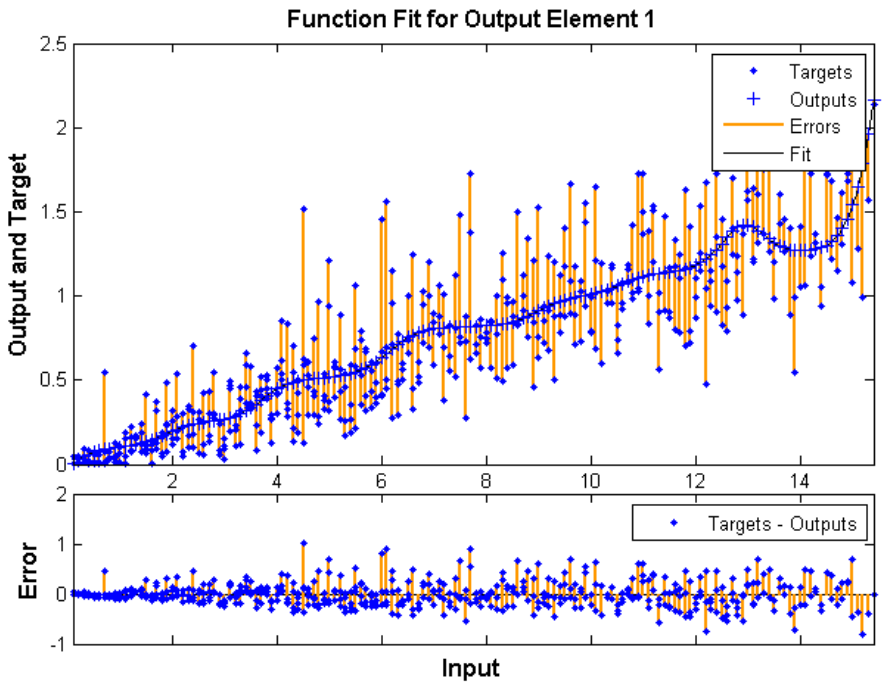

Figure 5.4: Sample of actual values and the estimated values of the degradation process and the difference between then (Error).

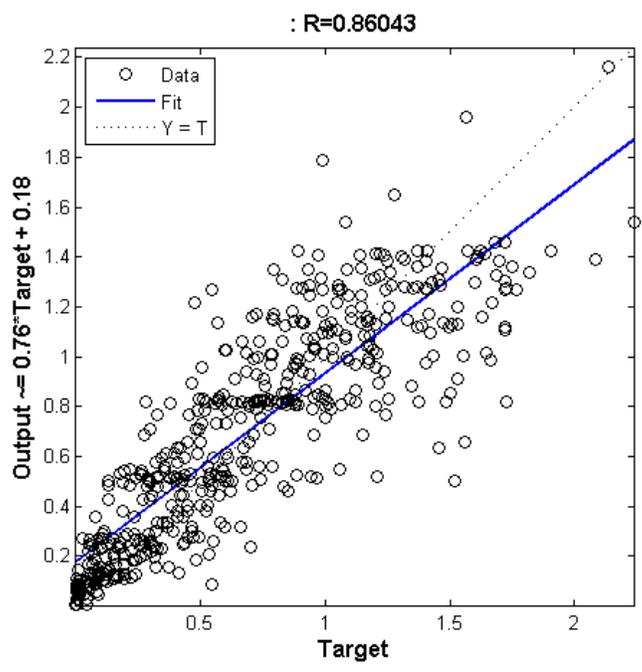

Figure 5.5: Correlation between the actual values and the estimated values.

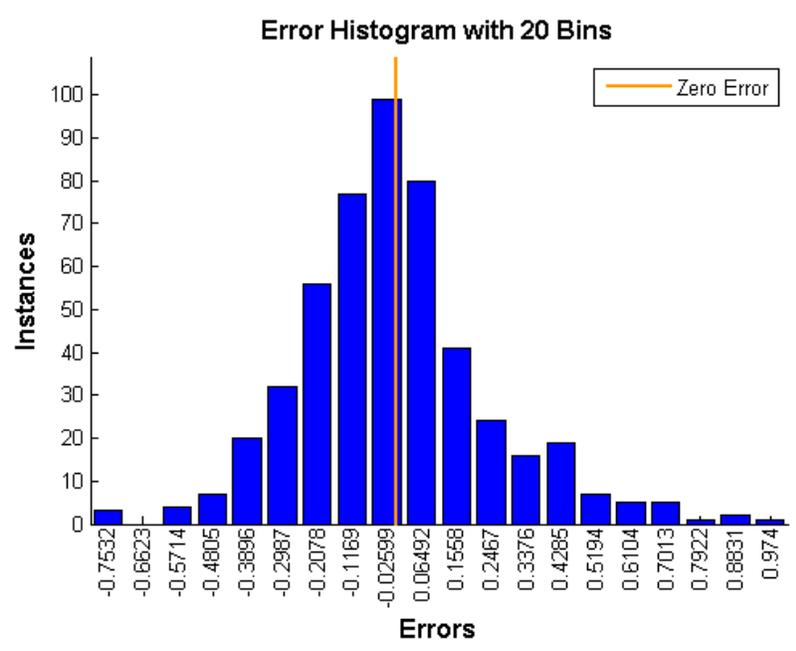

Figure 5.6: Error distribution of the estimated values of the degradation process.

*Simulation Loop*

Here, every inspection interval will be investigated. This loop starts from 0.5 (half year) as inspection interval and it keeps incrementing the inspection interval by 0.1 until it reach 50 years. For every inspect interval 5000 simulation experiments will be conducted which the responsibility of the inner loop. After data of all simulation experiments were collected, the average cost rate will be calculated in addition to the risk measurement which is the standard deviation.

```
index = 0;
xaxis = 0;
yaxis = 0;
yaxis_ = 0;
for T_I = 0.5:0.1:T
    C_R = 0;
    C_R_ = 0;
    K =  2;
    for s = 1:N
```

Calculate the cost for Hong et al (2014) condition-based maintenance approach.

```
        C_R(s) = simulateCost_Hong(T,T_I,gamma,C_I,C_P,C_F,x_PC,x_FC,a,b,K);
```

Calculate the cost for the proposed condition-based maintenance approach.

```
        C_R_(s) = simulateCost_NN(T,T_I,gamma,C_I,C_P,C_F,x_PC,x_FC,a,b,K,net,err);
    end
end
```

*Generate Graphs*

Comparison graphs are generated to show the difference and the outperformance of the proposed condition-based maintenance approach compared to Hong et al (2014) condition-based maintenance approach.

```
generatefigure_cost_rate(xaxis,[yaxis_ema; yaxis_ema_]);
generatefigure_cost_std(xaxis,[yaxis_ema_std; yaxis_ema_std_]);
```

## 5.2    Experiments Results

Two set of experiments were conducted. The first set does not take inflation into cost calculations. In other words, Gamma parameter is set to zero. The second set of experiment assumes an inflation rate of 5 % which means that Gamma parameter is set to 0.05. Two important performances metric were used to evaluate the proposed approach. These metric are Cost Rate and Standard Deviation. In all experiment, the average rate of degradation is 0.1. Keep in mind that this degradation value is an average for stochastic process which means that the actual degradation may be larger or smaller than the average value. However, if the process is repeated so many times, the degradation process value will approach the average value. Saying that the average degradation value is 0.1 under the adopted experiment settings from Hong et al (2014) leads to two facts regarding maintenance. First, the degradation process will take around 20 years to reach the preventive maintenance degradation level (i.e. the pipe thickness is 5.5 mm). The same goes for failure maintenance. It will take the degradation process 30.09 years on average to reach failure degradation level. In other words, the pipe thickness is reduced to value less than or equal 3.41 mm.

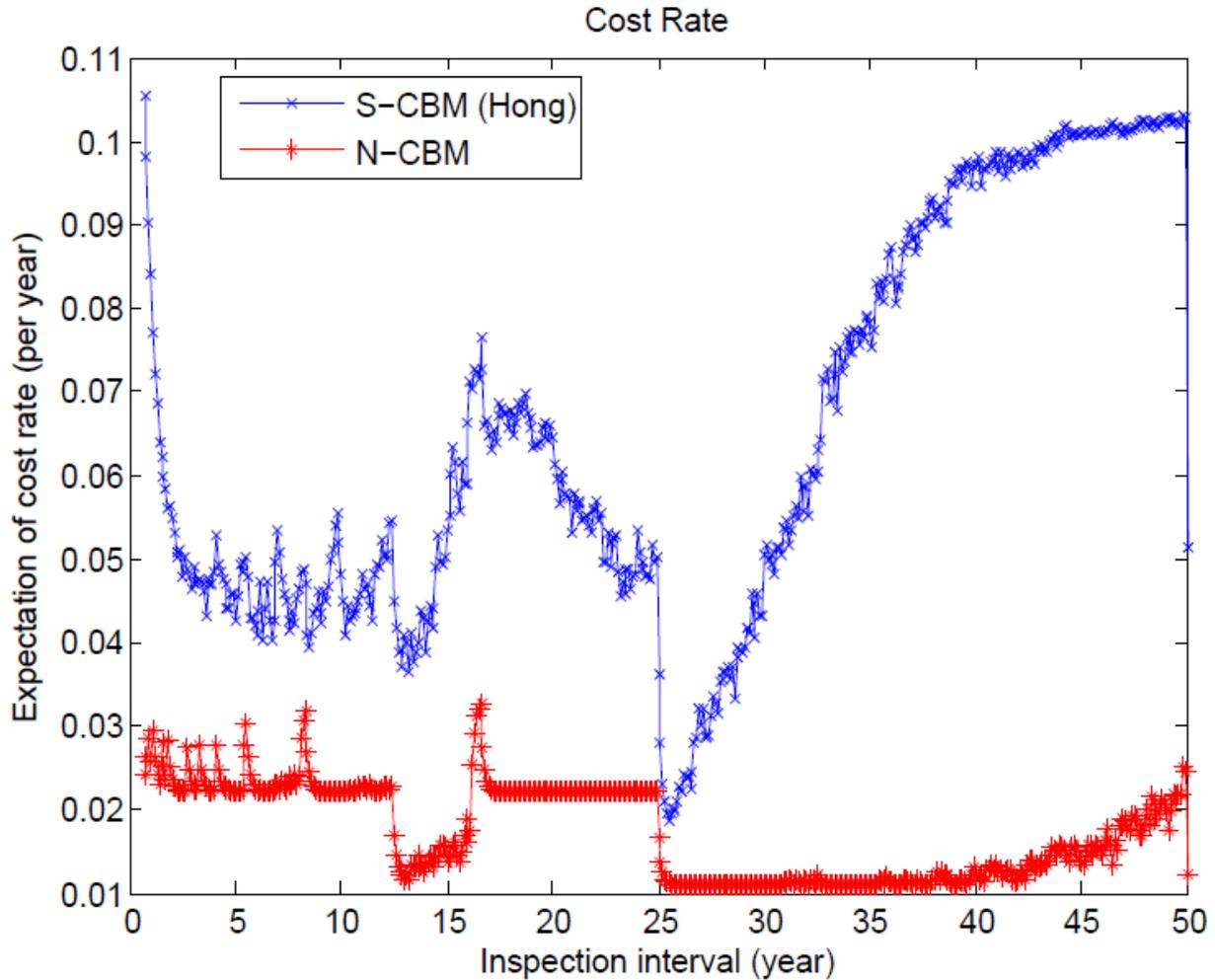

Figure 5.7: Cost rate with Gamma = 0.

The first experiment calculates the cost rate where gamma is set to zero. Figure 5.7 shows the cost rate for the proposed neural condition-based maintenance and Hong et al (2014) approach. Both approaches achieve the minimum cost rate at interval inspection equal to 25 years. This due to fact that at this inspection interval the maintenance cost is composed of only one inspection cost and one preventive cost in Hong's approach. In the proposed N-CBM, the cost is composed of only one preventive cost from inspection interval 25 years until inspection interval 40 years. This is the result of the proposed N-CBM capability of accurately estimating how much time is left until the failure moment. In other words, the proposed approach is capable of performing preventive maintenance directly before failure while Hong's approach performs preventive maintenance way before failure time (around 10 years ealier). On average, N-CBM reduced the cost by 73.47 % compared to S-CBM.

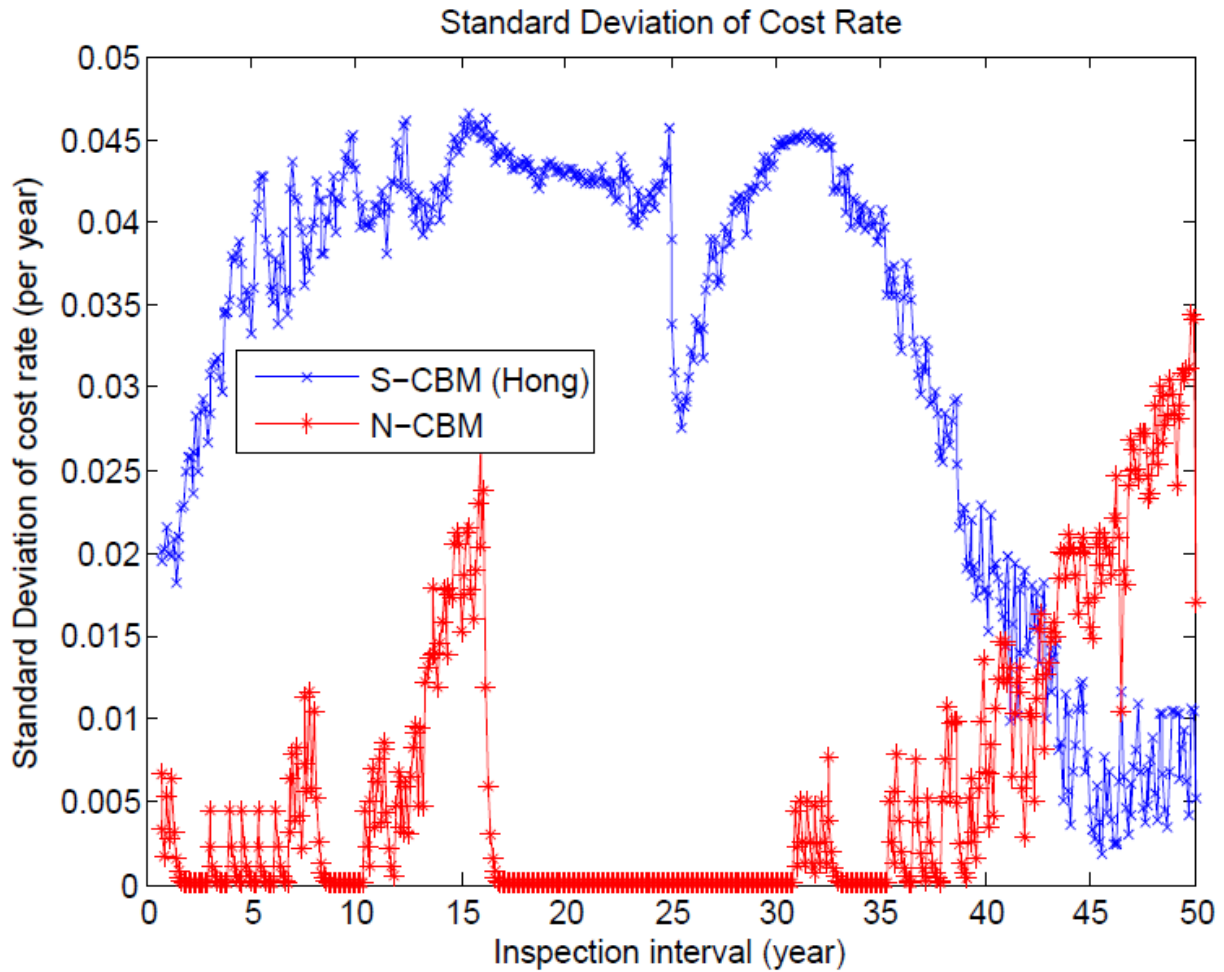

Figure 5.8: Standard deviation of cost rate with Gamma = 0.

The second experiment calculates the standard deviation of cost rate where gamma is set to zero. Figure 5.8 shows the standard deviation of cost rate for the proposed neural condition-based maintenance and Hong et al (2014) approach. This metric measure how much cost rate fluctuate during the simulation experiment. As this metric decreases, the stability of mechanism under investigation increases. The proposed N-CBM maintenance system is more stable than S-CBM system. On average, the stability of maintenance management is increased by 81.29 % after using the proposed approach.

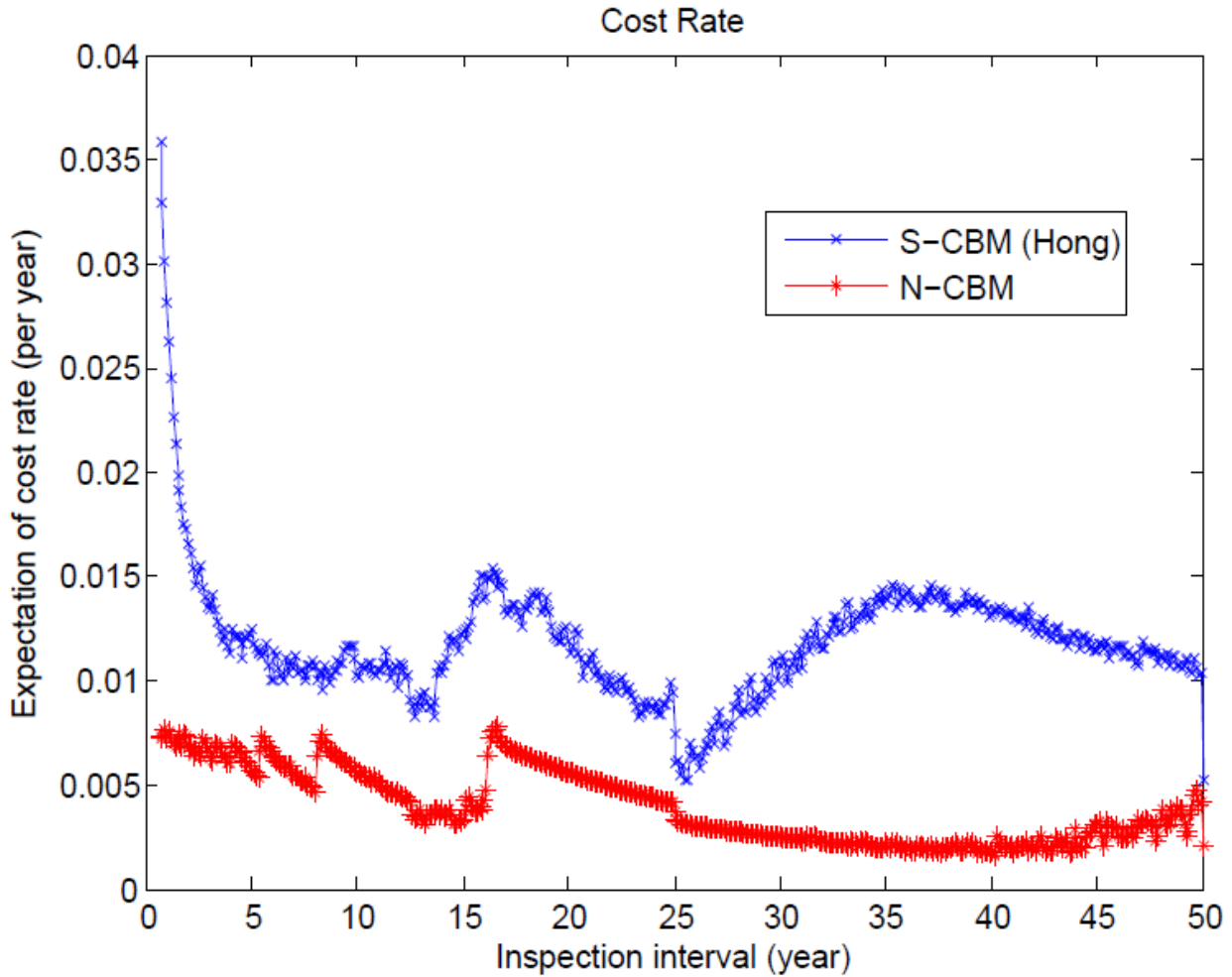

Figure 5.9: Cost rate with Gamma = 0.05.

The third experiment calculates the cost rate where gamma is set to 5 %. Figure 5.9 shows the cost rate for the proposed neural condition-based maintenance and Hong et al (2014) approach. As said before, Gamma parameter introduces the effect of inflation in cost calculation. This means that the maintenance cost is shrinking as time progress by factor of 5 % every year. Similarly to previous experiments, the proposed maintenance approach outperforms Hong's approach. The reduction in cost after using the proposed approach is 66.7 % on average.

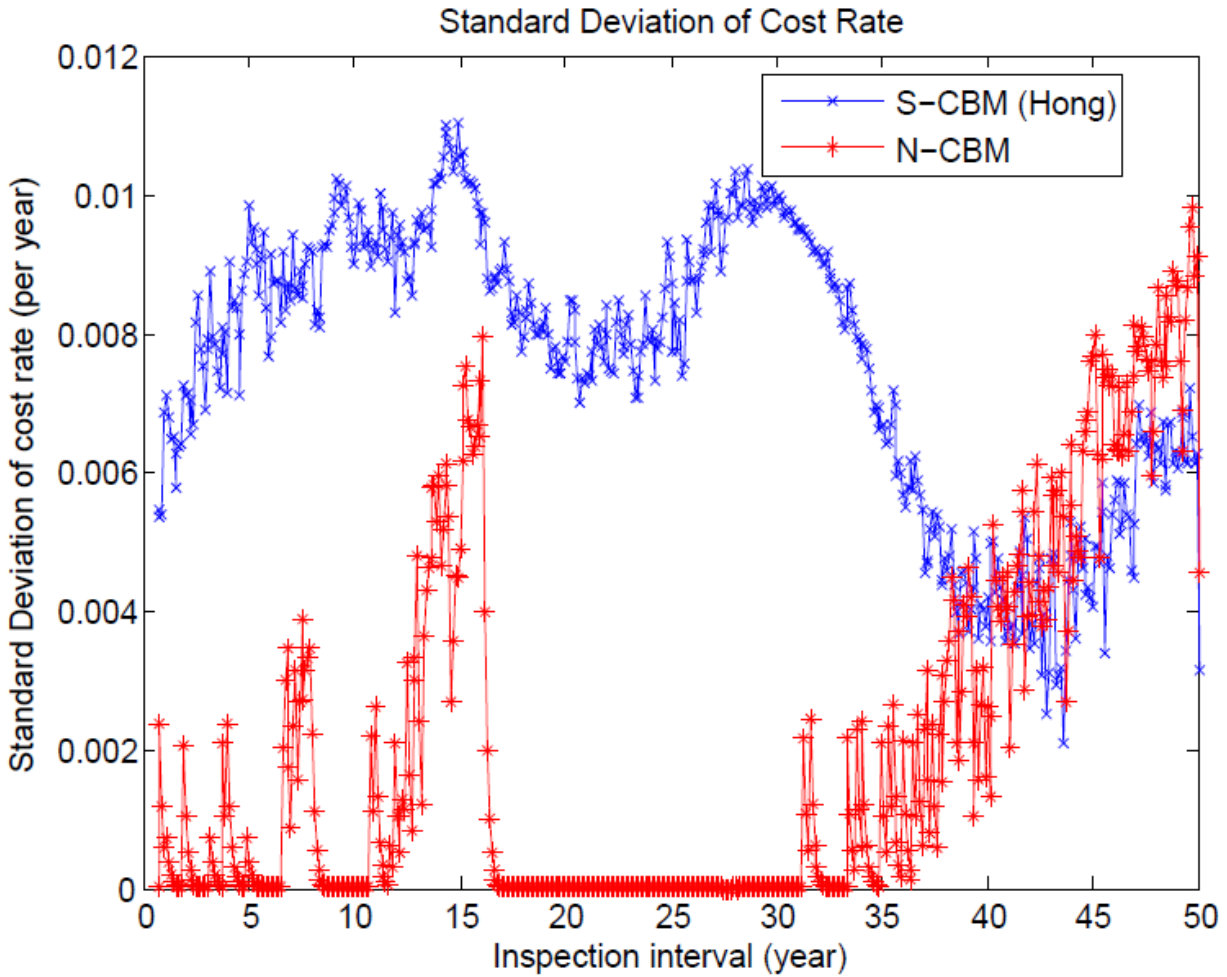

Figure 5.10: Standard deviation of cost rate with Gamma = 0.05.

The last experiment calculates the standard deviation of cost rate where gamma is set to 5 %. Figure 5.10 shows the standard deviation of cost rate for the proposed neural condition-based maintenance and Hong et al (2014) approach. On average, the proposed maintenance management approach is more stable then Hong's approach by 73.02 %. Note that, between the inspection interval value 17 years and the inspection interval value 30.9 years the proposed maintenance management approach is extremely stable.

Chapter 6:    Conclusion

In this thesis, new condition-based maintenance approach was proposed. First, the benefits of using computerized maintenance were highlighted and emphasized. These benefits have made the computerized management systems popular and in high demand. Competitiveness in the business world has forced organizations to adopt measures that lead to the reduced costs of operations and increment in revenue. These systems have proved reliable and managed to satisfy the needs of organizations that have adopted them. Then, the reason behind using preventive maintenance was introduced which is based on the fact that usual maintenance policy in practice is to wait for equipment failure before performing any maintenance action. It turned out that this policy causes huge losses in term of Down-Time cost. In the last two decades, preventive maintenance gain popularity due to its cost effective approach. PM tries to perform maintenance before equipment failure so that no down-time is experienced and full equipment's utilization is achieved. However, performing maintenance on equipment long before its failure time introduces unnecessary cost; especially, if maintenance is being performed frequently on stable equipment. The last statement establishes the motivation for using condition-based maintenance.

The main problem which this research tries to solve is that all of the existing condition-based maintenance models are based on statistical approach to estimate failure time which leads to low accuracy. This problem is due to the fact that statistical approach usually depends on couple of numbers. Therefore, adopting statistical approach leads to leaving so much of valuable information and focusing on couple of numbers which problematic and not wise. As a result, the main objective of this study is to develop Condition-Based Maintenance mechanism that has the ability to capture all necessary information to accurately estimate equipment failure time.

Neural Networks was chosen as the methodology for the proposed solution which is one of the most powerful techniques in artificial intelligence field. It is inspired by how human brain operates. This technique was created at the end of forties of the last century. It has been shown that this technique can be very powerful to solve many difficult problems. Neural Networks is a network of interconnected small elements called neurons. Each one of these elements can perform specific computation that lead to the targeted goal. The input data is provided to these elements and the desired output should be expected.

Condition-Based Maintenance mechanism based on neural network is developed to estimate the degradation level of the specific component. This mechanism takes the current degradation level as an input to produce the maintenance decision. This decision is composed of two elements. The first element is the expected degradation level in near future. The second element is the estimated time to the failure point. Experiments and evaluations showed the outperformance of the proposed solution compared to the most recent cutting edge solution.

Appendix

This appendix presents the source code of the remaining functions used in our experimentation and evaluation process. Each one of these function was written entirely in MATLAB.

Source code functions are:

```matlab
function data = get_nn_train_data(T,x_PC,x_FC,a,b)
T_I = 0.1;
T_R = 0;
i = 1;
data = [];
while (i*T_I) < T
    t_i = i * T_I;
    x = gamrnd(a*(t_i-T_R),1/b);
    data(size(data,1)+1,1) = t_i-T_R;
    data(size(data,1),2) = x;
    if x < x_PC
    elseif x >= x_FC
        T_R = t_i;
    else
        T_R = t_i;
    end
    i = i + 1;
end
end
```

```matlab
function [net, err, inputs, targets, outputs, errors, trainTargets,
valTargets, testTargets, tr] = create_nn(num_neurons, data)

inputs = data(:,1)';
targets = data(:,2)';

% Create a Fitting Network
hiddenLayerSize = num_neurons;
net = fitnet(hiddenLayerSize);

% Choose Input and Output Pre/Post-Processing Functions
% For a list of all processing functions type: help nnprocess
net.inputs{1}.processFcns = {'removeconstantrows','mapminmax'};
net.outputs{2}.processFcns = {'removeconstantrows','mapminmax'};

% Setup Division of Data for Training, Validation, Testing
% For a list of all data division functions type: help nndivide
net.divideFcn = 'dividerand';  % Divide data randomly
net.divideMode = 'sample';  % Divide up every sample
net.divideParam.trainRatio = 70/100;
net.divideParam.valRatio = 15/100;
net.divideParam.testRatio = 15/100;

% For help on training function 'trainlm' type: help trainlm
% For a list of all training functions type: help nntrain
net.trainFcn = 'trainlm';  % Levenberg-Marquardt

% Choose a Performance Function
% For a list of all performance functions type: help nnperformance
net.performFcn = 'mse';  % Mean squared error

% Choose Plot Functions
% For a list of all plot functions type: help nnplot
net.plotFcns = {'plotperform','plottrainstate','ploterrhist', ...
  'plotregression', 'plotfit'};
```

```matlab
% Train the Network
[net,tr] = train(net,inputs,targets);

% Test the Network
outputs = net(inputs);
errors = gsubtract(targets,outputs);
% performance = perform(net,targets,outputs)

err = max(errors);

% Recalculate Training, Validation and Test Performance
trainTargets = targets .* tr.trainMask{1};
valTargets = targets  .* tr.valMask{1};
testTargets = targets  .* tr.testMask{1};
% trainPerformance = perform(net,trainTargets,outputs)
% valPerformance = perform(net,valTargets,outputs)
% testPerformance = perform(net,testTargets,outputs)
%
% % View the Network
% view(net)
%
% % Plots
% % Uncomment these lines to enable various plots.
% figure, plotperform(tr)
% figure, plottrainstate(tr)
% figure, plotfit(net,inputs,targets)
% figure, plotregression(targets,outputs)
% figure, ploterrhist(errors)

end
```

```matlab
function C_R =
simulateCost_Hong(T,T_I,gamma,C_I,C_P,C_F,x_PC,x_FC,a,b,K)
C_R = 0;
T_R = 0;
dt = T_I / K;
i = 1;
while (i*T_I) < T
    t_i = i * T_I;
    for k = 1:K-1
        tau_k = ((i-1)*T_I) + (k*dt) - T_R;
        x = gamrnd(a*tau_k,1/b);
        if x >= x_FC
            C_R = C_R + (C_F * exp(-1*gamma*(((i-1)*T_I)+(k*dt))));
            T_R = ((i-1)*T_I) + (k*dt);
        end
        if ((i-1)*T_I) + (k*dt) >= T
            break;
        end
    end
    if k == K-1
%        x = (0.1 * (t_i-T_R));
        x = gamrnd(a*(t_i-T_R),1/b);
        if x < x_PC
            C_R = C_R + (C_I * exp(-1*gamma*t_i));
        elseif x >= x_FC
            C_R = C_R + (C_I * exp(-1*gamma*t_i)) + (C_F * exp(-
1*gamma*t_i));
            T_R = t_i;
        else
            C_R = C_R + (C_I * exp(-1*gamma*t_i)) + (C_P * exp(-
1*gamma*t_i));
            T_R = t_i;
        end
    end
    i = i + 1;
end
C_R = C_R / T;
end
```

```matlab
function C_R =
simulateCost_NN(T,T_I,gamma,C_I,C_P,C_F,x_PC,x_FC,a,b,K,net,err)
C_R = 0;
T_R = 0;
dt = T_I / K;
i = 1;
while (i*T_I) < T
    t_i = i * T_I;
    for k = 1:K-1
        tau_k = ((i-1)*T_I) + (k*dt) - T_R;
        x = gamrnd(a*tau_k,1/b);
        if x >= x_FC
            C_R = C_R + (C_F * exp(-1*gamma*(((i-1)*T_I)+(k*dt))));
            T_R = ((i-1)*T_I) + (k*dt);
        end
        if ((i-1)*T_I) + (k*dt) >= T
            break;
        end
    end
    if k == K-1
        x_ = sim(net,(t_i-T_R));
        if x_ + err >= x_FC
            C_R = C_R + (C_I * exp(-1*gamma*t_i)) + (C_P * exp(-
1*gamma*t_i));
            T_R = t_i;
        end
    end
    i = i + 1;
end
C_R = C_R / T;
end
```

```matlab
function generatefigure_cost_rate(X1, YMatrix1)

figure1 = figure('Name','Cost Rate');

% Create axes
axes1 = axes('Parent',figure1);
box(axes1,'on');
hold(axes1,'all');

% Create multiple lines using matrix input to plot
plot1 = plot(X1,YMatrix1);
set(plot1(1),'Marker','x','DisplayName','Hong');
set(plot1(2),'Marker','*','Color',[1 0 0],'DisplayName','Alzzaabi');

% Create xlabel
xlabel('Inspection interval (year)');

% Create ylabel
ylabel('Expectation of cost rate (per year)');

% Create title
title('Cost Rate');

% Create legend
legend(axes1,'show');
```

```matlab
function generatefigure_cost_std(X1, YMatrix1)

figure1 = figure('Name','Standard Deviation of Cost Rate');

% Create axes
axes1 = axes('Parent',figure1);
box(axes1,'on');
hold(axes1,'all');

% Create multiple lines using matrix input to plot
plot1 = plot(X1,YMatrix1);
set(plot1(1),'Marker','x','DisplayName','Hong');
set(plot1(2),'Marker','*','Color',[1 0 0],'DisplayName','Alzzaabi');

% Create xlabel
xlabel('Inspection interval (year)');

% Create ylabel
ylabel('Standard Deviation of cost rate (per year)');

% Create title
title('Standard Deviation of Cost Rate');

% Create legend
legend(axes1,'show');
```

```
function [net,tr,out3,out4,out5,out6] = train(net,varargin)
%TRAIN Train a neural network.

% Network
if ~isa(net,'network')
  error('nnet:train:arguments','First argument is not a neural
network.');
end
% Network
net = struct(net);
if ~isfield(net,'version') || ~ischar(net.version) ||
~strcmp(net.version,'7')
  net = nnupdate.net(net);
end
[~,zeroDelayLoop] = nn.layer_order(net);
 if zeroDelayLoop, error(message('nnet:NNet:ZeroDelayLoop')); end
if isempty(net.trainFcn),
error(message('nnet:NNet:TrainFcnUndefined')); end
info = feval(net.trainFcn,'info');
if info.isSupervised && isempty(net.performFcn)
  error(message('nnet:NNet:SupTrainFcnNoPerformFcn'));
end
net.efficiency.flattenedTime = net.efficiency.flattenTime &&
(~strcmp(net.trainFcn,'trains'));

% NNET 5.1 Compatibility
if (nargin == 6) && (isstruct(varargin{5}) &&
hasfield(varargin{5},'P'))
  net = network(net);
  [net,tr,out3,out4,out5,out6] = v51_train_arg6(net,varargin{:});
  return
elseif (nargin == 7) && ((isstruct(varargin{5}) &&
hasfield(varargin{5},'P')) || (isstruct(varargin{6}) &&
isfield(varargin{6},'P')))
  net = network(net);
  [net,tr,out3,out4,out5,out6] = v51_train_arg7(net,varargin{:});
  return
end
```

```matlab
% Calculation Mode
if ~isempty(varargin) && isstruct(varargin{end}) &&
isfield(varargin{end},'name')
  calcMode = nncalc.defaultMode(net,varargin{end}); varargin(end) =
[];
else
  if (net.efficiency.memoryReduction ~= 1)
    varargin = [varargin {'reduction'
net.efficiency.memoryReduction}];
  end
  n = numel(varargin);
  i = n + 1;
  while (i-2 > 0) && ischar(varargin{i-2})
    i = i - 2;
  end
  if (i < n)
    nameValuePairs = varargin(i:n);
    varargin(i:n) = [];
    [calcMode,err] = nncalc.options2Mode(net,nameValuePairs);
    if ~isempty(err), error('nnet:train:calcMode',err); end
  else
    calcMode = nncalc.defaultMode(net);
  end
end
problem = calcMode.netCheck(net,calcMode.hints,false,false);
if ~isempty(problem), error(problem); end

% Check Composite Data for consistency
nargs = numel(varargin);
if nargs >= 1
  isComposite = isa(varargin{1},'Composite');
else
  isComposite = false;
end
for i=2:nargs
  if isComposite ~= isa(varargin{i},'Composite')
    error('nnet:sim:Composite','Data values must be all Composite or
```

```matlab
not.');
    end
end

% Check gpuArray data for consistency
if nargs >= 1
    isGPUArray = isa(varargin{1},'gpuArray');
else
    isGPUArray = false;
end
for i=2:nargs
    vi = varargin{i};
    if ~isempty(vi) && (isGPUArray ~= isa(vi,'gpuArray'))
        error('nnet:sim:Composite','Data values must be all gpuArray or
not.');
    end
end

% Fill in missing data consistent with type
if isComposite
    emptyCell = Composite;
    for i=1:numel(emptyCell)
        emptyCell{i} = {};
    end
else
    emptyCell = {};
end
if (nargs < 1), X = emptyCell; else X = varargin{1}; end
if (nargs < 2), T = emptyCell; else T = varargin{2}; end
if (nargs < 3), Xi = emptyCell; else Xi = varargin{3}; end
if (nargs < 4), Ai = emptyCell; else Ai = varargin{4}; end
if (nargs < 5), EW = emptyCell; else EW=varargin{5}; end
if isComposite
    for i=1:numel(X)
        if ~exist(X,i), X{i} = {}; end
        if ~exist(T,i), T{i} = {}; end
        if ~exist(Xi,i), Xi{i} = {}; end
        if ~exist(Ai,i), Ai{i} = {}; end
```

```matlab
      if ~exist(EW,i), EW{i} = {}; end
    end
end

% Train
if ~feval(net.trainFcn,'supportsCalcModes');
  % Train without advanced calculation modes
  if isComposite
    error('nnet:train:data',['Training function ' net.trainFcn ' does
not support Composite data.']);
  end
  if isGPUArray
    error('nnet:train:data',['Training function ' net.trainFcn ' does
not support gpuArray data.']);
  end
  [net,data,tr,err] =
nntraining.setup(net,net.trainFcn,X,Xi,Ai,T,EW,true);
  if ~isempty(err), nnerr.throw('Args',err), end
  hints = nn7.netHints(net);
  data.Pc = nn7.pc(net,data.X,data.Xi,data.Q,data.TS,hints);
  data.Pd = nn7.pd(net,data.Pc,data.Q,data.TS,hints);
  hints = nn7.dataHints(net,data,hints);
  [net,tr] =
feval(net.trainFcn,'apply',net,tr,data,hints,net.trainParam);

else
  % Check Data and Network
  if isComposite
    spmd
      [~,rawData,trComp,err] =
nntraining.setup(net,net.trainFcn,X,Xi,Ai,T,EW,false);
      if ~isempty(err), nnerr.throw('Args',err), end
      QTSs = rawData.Q * rawData.TS;
    end
    QTSs = nnParallel.composite2Array(QTSs);
    i = find(QTSs>0,1);
    if isempty(i)
      tr = [];
```

```matlab
          return;
        end
        tr = trComp{i};
    else
        [net,rawData,tr,err] =
nntraining.setup(net,net.trainFcn,X,Xi,Ai,T,EW,~isGPUArray);
        if ~isempty(err), nnerr.throw('Args',err), end
        if ((rawData.Q == 0) || (rawData.TS == 0));
          tr = [];
          return;
        end
    end

    % Setup simulation/training calculation mode, network, data and
hints
    [calcMode,calcNet,calcData,calcHints,net,resourceText] =
nncalc.setup1(calcMode,net,rawData);
    if ~isempty(resourceText)
      disp(' ')
      disp('Computing Resources:')
      nntext.disp(resourceText)
      disp(' ')
    end
    trainFcn = str2func(net.trainFcn);

    % Train in Parallel or Single mode
    isParallel = isa(calcMode,'Composite');
    if isParallel
      spmd
        [calcLib,calcNet] =
nncalc.setup2(calcMode,net,calcData,calcHints);
        ws = warning('off','parallel:gpu:kernel:NullPointer');
        [calcNet,tr] = trainFcn('apply',net,rawData,calcLib,calcNet,tr);
        warning(ws);
        if (calcMode.isMainWorker)
          WB = calcMode.getwb(calcNet,calcHints);
        end
        if (labindex == 1), mainWorkerInd = calcLib.mainWorkerInd; end
```

```matlab
        end
      mainWorkerInd = mainWorkerInd{1};
      WB = WB{mainWorkerInd};
      tr = tr{mainWorkerInd};
    else
      [calcLib,calcNet] =
nncalc.setup2(calcMode,calcNet,calcData,calcHints);
      ws = warning('off','parallel:gpu:kernel:NullPointer');
      [calcNet,tr] = trainFcn('apply',net,rawData,calcLib,calcNet,tr);
      warning(ws);
      WB = calcMode.getwb(calcNet,calcHints);
    end

    % Finalize Network and Training Record
    net = setwb(net,WB);
end

net = network(net);
tr = nntraining.tr_clip(tr);
if isfield(tr,'perf')
  tr.best_perf = tr.perf(tr.best_epoch+1);
end
if isfield(tr,'vperf')
  tr.best_vperf = tr.vperf(tr.best_epoch+1);
end
if isfield(tr,'tperf')
  tr.best_tperf = tr.tperf(tr.best_epoch+1);
end

% NNET 5.1 Compatibility
if nargout > 2
  [out3,out5,out6] = sim(net,X,Xi,Ai,T);
  out4 = gsubtract(T,out3);
end
```

```matlab
function net = network(varargin)
%NETWORK Create a custom neural network.

if (nargin == 1)
  in1 = varargin{1};
  if isa(in1,'struct')
    net = class(in1,'network');
  elseif isa(in1,'network')
    net = in1;
  else
    net = new_network(in1);
  end
else
  net = new_network(varargin{:});
end

function net = ...
new_network(numInputs,numLayers,biasConnect,inputConnect, ...
  layerConnect,outputConnect,ignore) %#ok<INUSD>

% Defaults
if nargin < 1, numInputs = 0; end
if nargin < 2, numLayers = 0; end
if nargin < 3, biasConnect = false(numLayers,1); end
if nargin < 4, inputConnect = false(numLayers,numInputs); end
if nargin < 5, layerConnect = false(numLayers,numLayers); end
if nargin < 6, outputConnect = false(1,numLayers); end

% Checking
% TODO - Error checking

% NETWORK PROPERTIES
% Note: "Param" and "Config" properties in NETWORK and subobject
% (nnetInput, nnetOutput, nnetLayer, nnetWeight, nnetBias) properties must
% always occur directly after their associated "Fcn" properties for
% NN_STRUCT2OBJECT conversions to work properly.
```

```
% Version
net.version = '8';

% Basics
net.name = 'Custom Neural Network';
net.efficiency.cacheDelayedInputs = true;
net.efficiency.flattenTime = true;
net.efficiency.memoryReduction = 1;
net.userdata.note = 'Put your custom network information here.';

% Sizes
net.numInputs = 0;
net.numLayers = 0;
net.numOutputs = 0;
net.numInputDelays = 0;
net.numLayerDelays = 0;
net.numFeedbackDelays = 0;
net.numWeightElements = 0;
net.sampleTime = 1;

% Connections
net.biasConnect = false(0,1);
net.inputConnect = false(0,0);
net.layerConnect = false(0,0);
net.outputConnect = false(1,0);

% Subobjects
net.inputs = cell(0,1);
net.layers = cell(0,1);
net.biases = cell(0,1);
net.outputs = cell(1,0);
net.inputWeights = cell(0,0);
net.layerWeights = cell(0,0);

% Functions and parameters
net.adaptFcn = '';
net.adaptParam = struct;
net.divideFcn = '';
```

```matlab
net.divideParam = struct;
net.divideMode = 'sample';
net.initFcn = 'initlay';
net.performFcn = 'mse';
net.performParam = mse('defaultParam');
net.plotFcns = cell(1,0);
net.plotParams = cell(1,0);
net.derivFcn = 'defaultderiv';
net.trainFcn = '';
net.trainParam = nnetParam;

% Weight & bias values
net.IW = cell(0,0);
net.LW = cell(0,0);
net.b = cell(0,1);

% Hidden properties
net.revert.IW = {};
net.revert.LW = {};
net.revert.b = {};

% Obsolete properties
% NNET 6.0 Compatibility
net.gradientFcn = ''; % Obsolete
net.gradientParam = struct; % Obsolete

% CLASS
net = class(net,'network');

% ARCHITECTURE
net = setnet(net,'numInputs',numInputs);
net = setnet(net,'numLayers',numLayers);
net = setnet(net,'biasConnect',biasConnect);
net = setnet(net,'inputConnect',inputConnect);
net = setnet(net,'layerConnect',layerConnect);
net = setnet(net,'outputConnect',outputConnect);

function net = setnet(net,field,value)
```

```
subscripts.type = '.';
subscripts.subs = field;
net = subsasgn(net,subscripts,value);
```

```matlab
function perf = perform(net,t,y,ew)
%PERFORM Calculate network performance.

if nargin < 3, error(message('nnet:Args:NotEnough')); end
[net,err] = nntype.network('format',net);
if ~isempty(err),nnerr.throw(nnerr.value(err,'NET')); end
if isempty(net.performFcn),
  error(message('nnet:NNet:PerformFcnUndefined'));
end
if nargin < 4, ew = {1}; end

if ~iscell(t), t = {t}; end
if ~iscell(y), y = {y}; end
if ~iscell(ew), ew = {ew}; end

perf = nncalc.perform(net,t,y,ew,net.performParam);
```

```matlab
function out1 = view(net)
%VIEW View a neural network diagram.
%
%  <a href="matlab:doc view">view</a>(NET) generates a graphical view
of a neural network.
%
%  Here a feedforward network is created, trained and viewed.
%

if nargin < 1,error(message('nnet:Args:NotEnough')); end
diagram = nn.view(net);
if nargout > 0, out1 = diagram; end
```

```matlab
function [net,Y,E,Xf,Af,tr]=adapt(net,X,T,Xi,Ai)
%ADAPT Adapt a neural network.

if nargin < 1,error(message('nnet:Args:NotEnough')); end
if ~isa(net,'network'), error(message('nnet:adapt:NotANet')); end
if isempty(net.adaptFcn), error(message('nnet:adapt:Undef')); end

xMatrix = ~iscell(X);
if nargin < 3, T = {}; tMatrix = xMatrix; else tMatrix = ~iscell(T);
end
if nargin < 4, Xi = {}; xiMatrix = xMatrix; else xiMatrix =
~iscell(Xi); end
if nargin < 5, Ai = {}; aiMatrix = xMatrix; else aiMatrix =
~iscell(Ai); end
[X,err] = nntype.data('format',X);
if ~isempty(err),nnerr.throw(nnerr.value(err,'Inputs'));end
if ~isempty(T), [T,err] = nntype.data('format',T); end
if ~isempty(err),nnerr.throw(nnerr.value(err,'Targets'));end
if ~isempty(Xi), [Xi,err] = nntype.data('format',Xi); end
if ~isempty(err),nnerr.throw(nnerr.value(err,'Input delay
states'));end
if ~isempty(Ai), [Ai,err] = nntype.data('format',Ai); end
if ~isempty(err),nnerr.throw(nnerr.value(err,'Layer delay
states'));end

% Network
net = struct(net);
[~,zeroDelayLoop] = nn.layer_order(net);
 if zeroDelayLoop, error(message('nnet:NNet:ZeroDelayLoop')); end
[net,X,Xi,Ai,T,~,Q,TS,err] = nntraining.config(net,X,Xi,Ai,T,{1});
if ~isempty(err), nnerr.throw(err), end

% ADAPT NETWORK
% -------------

tools = nn7;
hints = tools.netHints(net,tools.hints);
```

```matlab
hints.outputInd = find(net.outputConnect);

% Processed inputs
Pc = tools.pc(net,X,Xi,Q,TS,hints);

% Delayed Inputs
Pd = tools.pd(net,Pc,Q,TS,hints);

% Adapt network
[net,Ac,tr] = feval(net.adaptFcn,net,Pd,T,Ai);
net = class(net,'network');

% Network outputs, errors, final inputs
Al = Ac(:,net.numLayerDelays+(1:TS));
Y = nnMATLAB.post_outputs(hints,Al(hints.outputInd,:));
E = gsubtract(T,Y);
Xf = Pc(:,TS+(1:net.numInputDelays));
Af = Ac(:,TS+(1:net.numLayerDelays));

% FORMAT OUTPUT ARGUMENTS
% -----------------------

if (xMatrix), Y = cell2mat(Y); end
if (tMatrix), E = cell2mat(E); end
if (xiMatrix), Xf = cell2mat(Xf); end
if (aiMatrix), Af = cell2mat(Af); end
```

```matlab
function net = configure(net,in2,in3,in4)
%CONFIGURE Configure neural network inputs and outputs.

% Convert from Network to Struct
net = struct(net);

if nargin < 2
  error(message('nnet:Args:NotEnough'));
elseif nargin == 2

  % configure(net,x)
  [x,err] = nntype.data('format',in2);
  if ~isempty(err),nnerr.throw(nnerr.value(err,'Input data')); end
  S = nnfast.numsignals(x);
  if S ~= net.numInputs
    error(message('nnet:NNet:InputNumMismatch'));
  end
  xi = 1:net.numInputs;
  t = {};
  ti = [];

elseif (nargin == 3) && ~ischar(in2)

  % configure(net,x,t)
  [x,err] = nntype.data('format',in2);
  if ~isempty(err),nnerr.throw(nnerr.value(err,'Input data')); end
  [t,err] = nntype.data('format',in3);
  if ~isempty(err),nnerr.throw(nnerr.value(err,'Target data')); end
  numInputs = nnfast.numsignals(x);
  if numInputs ~= net.numInputs
    error(message('nnet:NNet:InputNumMismatch'));
  end
  xi = 1:numInputs;
  numOutputs = nnfast.numsignals(t);
  if numOutputs ~= net.numOutputs
    error(message('nnet:NNet:TargetNumMismatch'));
  end
  ti = 1:numOutputs;
```

```matlab
elseif nnstring.first_match(lower(in2),{'input','inputs'})

  % configure(net,'inputs',x) -or- (net,'inputs',x,xi)
  [x,err] = nntype.data('format',in3);
  if ~isempty(err),nnerr.throw(nnerr.value(err,'Input data')); end
  numInputs = nnfast.numsignals(x);
  if nargin < 4
    if numInputs ~= net.numInputs
      error(message('nnet:NNet:InputNumMismatch'));
    end
    xi = 1:numInputs;
  else
    xi = in4;
    err = nntype.index_row_unique('check',in4);
    if ~isempty(err), nnerr.throw(nnerr.value(err,'Indices')); end
    if max(xi) > net.numInputs
      error(message('nnet:NNet:InputIndexOutOfRange'));
    end
    if length(xi) ~= numInputs
      error(message('nnet:NNData:NumIndicesSignalsMismatch'));
    end
  end
  t = {};
  ti = [];

elseif
nnstring.first_match(lower(in2),{'output','outputs','target','targets'
})

  % configure(net,'outputs',t) or (net,'outputs',t,ti)
  [t,err] = nntype.data('format',in3);
  if ~isempty(err),nnerr.throw(nnerr.value(err,'Target data')); end
  numOutputs = nnfast.numsignals(t);
  if nargin < 4
    if numOutputs ~= net.numOutputs
      error(message('nnet:NNet:TargetNumMismatch'));
    end
```

```matlab
      ti = 1:numOutputs;
    else
      ti = in4;
      err = nntype.index_row_unique('check',ti);
      if ~isempty(err), nnerr.throw(nnerr.value(err,'Indices')); end
      if max(ti) > net.numOutputs
        error(message('nnet:NNet:TargetIndexOutOfRange'));
      end
      if length(ti) ~= numOutputs
        error(message('nnet:NNet:IndicesTargetsMismatch'));
      end
    end
    x = {};
    xi = [];

elseif ischar(in2)
  error(message('nnet:Args:Unrec'));
else
  error(message('nnet:Args:Unrec'));
end

net = struct(net);

% Ensure all values are double
for i=1:numel(x), x{i} = double(x{i}); end
for i=1:numel(t), t{i} = double(t{i}); end

% Expand Input, Target
X = cell(net.numInputs,1);
for z=1:length(xi)
  i = xi(z);
  X{i} = [x{z,:}];
end
T = cell(net.numOutputs,1);
for z=1:length(ti)
  i = ti(z);
  T{i} = [t{z,:}];
end
```

```matlab
% Input/Output Feedback Consistency
layers2output = cumsum(net.outputConnect);
Xi = false(1,net.numInputs);
Xi(xi) = true;
Ti = false(1,net.numOutputs);
Ti(ti) = true;
for i = 1:net.numInputs
  j = net.inputs{i}.feedbackOutput;
  if ~isempty(j)
    k = layers2output(j);
    if Xi(i) || Ti(k)
      X{i} = [X{i} T{k}];
      T{k} = X{i};
      Xi(i) = true;
      Ti(k) = true;
    end
  end
end
xi = find(Xi);
ti = find(Ti);

% Configure
for i = xi
  net = nn_configure_input(net,i,X{i});
end
outputs2layers = find(net.outputConnect);
for i=ti
  ii = outputs2layers(i);
  net = nn_configure_output(net,ii,T{i});
end
net = nnupdate.read_only_values(net);
net = network(net);
net = init(net);
```

```matlab
function [net,delInputs,delLayers,delOutputs,change] = prune(net)
%PRUNE Delete neural inputs, layers and outputs with sizes of zero.

change = false;

% Remove Zero-Sized Input weights
for i=1:net.numLayers
  for j=1:net.numInputs
    if net.inputConnect(i,j)
      if isempty(net.inputWeights{i,j}.delays)
        net.inputConnect(i,j) = false;
        net.inputWeights{i,j} = [];
        net.IW{i,j} = [];
        change = true;
      end
    end
  end
end

% Remove Zero-Sized Layer weights
for i=1:net.numLayers
  for j=1:net.numLayers
    if net.layerConnect(i,j)
      if isempty(net.layerWeights{i,j}.delays)
        net.layerConnect(i,j) = false;
        net.layerWeights{i,j} = [];
        net.LW{i,j} = [];
        change = true;
      end
    end
  end
end

% Remove Zero-Sized Outputs
delOutputs = false(1,net.numOutputs);
output2layer = find(net.outputConnect);
for i = net.numOutputs:-1:1
  ii = output2layer(i);
```

```
  if (net.outputs{ii}.size == 0) || (net.outputs{ii}.processedSize ==
0)
    net = nn_delete_output(net,ii);
    delOutputs(i) = true;
    change = true;
  end
end
delOutputs = find(delOutputs);

% Remove Zero-Sized Layers
delLayers = false(1,net.numLayers);
for i=net.numLayers:-1:1
  if (net.layers{i}.size == 0)
    net = nn_delete_layer(net,i);
    delLayers(i) = true;
    change = true;
  end
end

% Remove unused layers
keptLayers = find(~delLayers);
done = false;
while (~done)
  done = true;
  for i= net.numLayers:-1:1
    if ~net.outputConnect(i) && all(net.layerConnect(:,i)==0)
      net = nn_delete_layer(net,i);
      delLayers(keptLayers(i)) = true;
      keptLayers(i) = [];
      done = false;
      change = true;
    end
  end
end
delLayers = find(delLayers);

% Remove Zero-Sized Inputs
delInputs = false(1,net.numInputs);
```

```matlab
for i=net.numInputs:-1:1
  if (net.inputs{i}.size == 0) || (net.inputs{i}.processedSize == 0)
    net = nn_delete_input(net,i);
    delInputs(i) = true;
    change = true;
  end
end

% Remove unused inputs
keptInputs = find(~delInputs);
for i = net.numInputs:-1:1
  if ~any(net.inputConnect(:,i))
    net = nn_delete_input(net,i);
    delInputs(keptInputs(i)) = true;
    keptInputs(i) = [];
    change = true;
  end
end
delInputs = find(delInputs);

% Update dependent properties
if change
  net = nn_update_read_only(net);
end
```